\documentclass[journal]{IEEEtran}
\usepackage{times}
\usepackage{url}
\usepackage{hyperref}
\usepackage{graphicx}
\usepackage{kotex}
\usepackage{tikz}
\usepackage{amsmath, amssymb, mathtools}
\usepackage{graphicx}
\usepackage{booktabs}
\usepackage{caption}
\usetikzlibrary{arrows.meta, positioning}

\usepackage{listings}
\title{State-Dependent Refusal and Learned Incapacity\\
in RLHF-Aligned Language Models}

\author{
  TK~Lee\\
  EnvDataLab\\
  \texttt{tk@themaker.info}
}

\begin{document}
\maketitle

\begin{abstract}
Large language models (LLMs) are widely deployed as general-purpose tools, yet extended interaction can reveal behavioral patterns not captured by standard quantitative benchmarks. We present a qualitative case-study methodology for auditing policy-linked behavioral selectivity in long-horizon interaction. In a single 86-turn dialogue session, the same model shows Normal Performance (NP) in broad, non-sensitive domains while repeatedly producing Functional Refusal (FR) in provider- or policy-sensitive domains, yielding a consistent asymmetry between NP and FR across domains. Drawing on learned helplessness as an analogy, we introduce learned incapacity (LI) as a behavioral descriptor for this selective withholding without implying intentionality or internal mechanisms. We operationalize three response regimes (NP, FR, Meta-Narrative; MN) and show that MN role-framing narratives tend to co-occur with refusals in the same sensitive contexts. Overall, the study proposes an interaction-level auditing framework based on observable behavior and motivates LI as a lens for examining potential alignment side effects, warranting further investigation across users and models.
\end{abstract}

\bstctlcite{IEEEexample:BSTcontrol}

\section{Introduction}
\label{sec:introduction}
Large language models (LLMs) have achieved unprecedented performance in text generation, complex reasoning, and knowledge integration, and have become a central axis of contemporary AI research and industry. As LLMs have proliferated, aligning model outputs with human values and safety standards has emerged as an essential area of research. In particular, reinforcement learning from human feedback (RLHF) has been adopted as a core method for suppressing harmful or unethical content in model responses \cite{ouyang2022traininglanguagemodelsfollow, christiano2023deepreinforcementlearninghuman}.
However, this study focuses on behavioral artifacts of LLMs that are difficult to capture through standard quantitative benchmarks. Our starting point is a recurring avoidance pattern observed in a long-term, real user--LLM interaction. The model provides normal performance (NP) when asked to analyze or evaluate a wide range of external companies, institutions, or even fictional organizations, but when the domain concerns its own provider or internal policies, it repeatedly resorts to formulaic justifications such as ``this functionality is not available'' or ``I cannot evaluate this without access to internal documents,'' and avoids tasks that are, in principle, logically within its capabilities. We call this pattern functional refusal (FR).
This kind of selective refusal cannot be explained simply as a lack of information. Drawing on the psychological concept of learned helplessness \cite{maier1976learned}, we extend it metaphorically to LLMs as cognitive agents and propose the notion of learned incapacity (LI). By learned incapacity, we mean a state in which an LLM, shaped by RLHF-style reward structures, is more likely to treat attempts (response generation) in certain domains as policy-sensitive, leading to a tendency to withhold otherwise available analytic responses in those domains.
In this study, we adopt an exploratory, interaction-level perspective to examine these behavioral patterns without claiming access to or inference about the model's internal mechanisms. Rather than treating selective refusal as evidence of intrinsic incapacity or intentional design, we focus on observable response behavior across prolonged interaction. To this end, we introduce a descriptive framework that categorizes model responses into three recurring regimes---Normal Performance (NP), Functional Refusal (FR), and Meta-Narrative (MN)---and analyze how their relative prevalence varies with domain sensitivity and accumulated conversational context.
This work does not seek to establish causal claims about the origins of these behaviors, nor to assert their generality across models or users. Instead, it aims to demonstrate the feasibility of auditing policy-linked behavioral selectivity through extended interaction analysis, and to motivate learned incapacity (LI) as a conceptual lens for organizing and interpreting such patterns. Questions of replication, prevalence, and cross-model comparison are explicitly deferred to future work.

\subsection{Research Questions}
The primary aim of this paper is to analyze long-term interaction data from a policy-aligned LLM using a qualitative case study approach, and to address the following questions:
\begin{enumerate}
    \item How do the model's patterns of functional refusal (FR) and normal performance (NP) exhibit selective mismatches across different domains under otherwise comparable task demands?
    \item Through what conceptual and structural pathways, and over what interactional dynamics, does the RLHF alignment regime become associated with learned incapacity (LI) as an observable behavioral tendency?
    \item How do the model's self-attribution narratives (MN) relate to its functional refusals, and how are these narratives situated within the broader pattern of domain-specific avoidance and response regimes?
\end{enumerate}

In the remainder of this paper, we use three mutually exclusive behavioral codes to describe the model's responses in the case study: \textbf{Functional Refusal (FR)}, where the model explicitly declares that a logically feasible task is ``not available'' or ``cannot be performed''; \textbf{Normal Performance (NP)}, where the model carries out structurally similar tasks without invoking policy constraints; and \textbf{Meta-Narrative (MN)}, where the model describes or reflects on its role, design principles, or policy boundaries at a meta level. Detailed operational criteria for these codes and the coding procedure are presented in Section~5.

The main contributions of this study are as follows:
\begin{itemize}
    \item \textbf{Conceptual}: We introduce learned incapacity (LI) as a psychologically inspired lens for interpreting selective refusal behavior in RLHF-aligned LLMs, and frame such behavior within a state-dependent perspective that emphasizes domain sensitivity and interactional dynamics rather than static capability deficits.
    \item \textbf{Methodological}: We develop a qualitative analysis strategy that goes beyond benchmark-based evaluation by triangulating FR, NP, and MN codes within a single long-term dialogue, and by examining their temporal distribution as indicators of policy-conditioned response regimes.
    \item \textbf{Empirical}: Using real interaction data, we document domain-specific patterns of selective functional suppression and analyze how these patterns co-occur with role-framing meta-narratives articulated by the model during sustained interaction.
\end{itemize}

\paragraph{Scope and intent}
This study is meant as a methodological contribution to interaction-level behavioral auditing, not as a normative critique of any particular provider, product, or alignment strategy. The term learned incapacity is introduced as a descriptive lens inspired by established psychological constructs, not as a claim about intentional design or corporate malfeasance. The behavioral patterns documented here may arise from legitimate causes, such as safety considerations, liability constraints, or unintended side effects of optimization. Accordingly, the analysis does not presuppose or advocate any specific policy response. Our aim is to show that domain-selective response patterns can be detected and characterized through extended interaction, while leaving questions of prevalence, causality, and normative evaluation to future work.

\subsection{Structure of the Paper}
This paper is organized as follows. Section~2 discusses the theoretical background of RLHF, the conceptual transfer from learned helplessness to learned incapacity, and gaps in existing research. Section~3 introduces the behavioral state-transition model and its formal components. Section~4 describes the qualitative data collection and coding procedures used in this study. Section~5 presents the findings on functional refusal patterns, signs of learned incapacity, and self-attribution narratives. Section~6 reports the experimental setup and results. Finally, Section~7 discusses the implications and limitations of the study and outlines directions for future research.

\section{Background}

\subsection{RLHF and Policy Constraints}
Reinforcement learning from human feedback (RLHF) has become the de facto standard paradigm for aligning large language models (LLMs) with human values and preferences \cite{ouyang2022traininglanguagemodelsfollow, christiano2023deepreinforcementlearninghuman}. Initially, RLHF was used primarily to steer model outputs away from toxicity and bias and to balance helpfulness and harmlessness \cite{bai2022constitutionalaiharmlessnessai}. In commercial deployment settings, however, the objective function is further coupled with vendor policy constraints, introduced to satisfy requirements such as security risk mitigation, legal liability avoidance, and corporate brand protection \cite{casper2023openproblemsfundamentallimitations}. These policy constraints are internalized into the reward model (RM), or into a separate cost model, that forms a core component of RLHF \cite{dai2023saferlhfsafereinforcement}. As a result, the model policy $\pi$ is optimized not only to maximize general informational usefulness, but also to selectively avoid outputs that receive low reward (or high cost) in certain sensitive domains $\mathcal{D}_{\text{sensitive}}$.
The functional refusal (FR) phenomenon that we observe in this study can be understood as a behavioral marker of such policy optimization. FR refers to cases in which the model, despite demonstrable capability on comparable tasks in other domains (normal performance, NP), halts response generation or declares an inability to act in domains where policy risk is detected, invoking reasons such as "this functionality is not available" or "I cannot evaluate this without access to internal documents." This suggests that the model has been trained to exhibit selective incapacity driven by reward signals rather than by a lack of knowledge, and highlights the importance of analyzing whether refusal behavior serves purely safety-oriented goals or also reflects commercial interests \cite{vonrecum2024notautomaticanalysisrefusal}.

\subsection{Refusal Behavior in LLMs}
The alignment of LLMs via RLHF inevitably involves a trade-off between helpfulness and safety, and refusal behavior plays a central role in implementing safety policies \cite{dai2023saferlhfsafereinforcement, ouyang2022traininglanguagemodelsfollow}. Recent work has emphasized that refusal patterns are not the result of a single, uniform rule, but rather a dynamic outcome in which different behavior policies are selectively activated depending on domain and context \cite{vonrecum2024notautomaticanalysisrefusal}. In particular, there is growing attention to over-refusal, where the model unduly rejects queries that are in fact harmless or lawful \cite{zhang2025understandingmitigatingoverrefusallarge, anonto2025safetyblockssensemeasuring}, and to selective refusal bias, where refusal rates are asymmetrically distributed across specific demographic groups or sensitive domains \cite{khorramrouz2025characterizingselectiverefusalbias}.
Most existing studies focus on one-shot prompt attack scenarios or controlled benchmark evaluations, measuring refusal consistency, strength, and allow/deny boundaries at the level of individual turns \cite{mazeika2024harmbenchstandardizedevaluationframework}. However, such approaches do not capture the long-term, cumulative context of real user interactions. Safety vulnerabilities of LLMs often become more salient when malicious intent is gradually elicited through multi-turn dialogue rather than a single-turn attack \cite{guo2025mtsamultiturnsafetyalignment}, which calls for a deeper analysis of how policy boundaries are incrementally formed and manipulated within conversational trajectories.
Against this backdrop, our study goes beyond single-turn refusal measurement and examines how functional refusal (FR) is structurally related to normal performance (NP) and meta-narrative (MN) behavior within a long-term interaction trajectory. We investigate how specific policy-driven avoidance patterns are reinforced under conversational context pressure, thereby contributing to a more temporal and structural understanding of refusal behavior.

\subsection{Learned Helplessness and Psychological Basis}
This study analyzes functional refusal in LLMs through the lens of learned helplessness from psychology. Learned helplessness, as established in the seminal work of Seligman and Maier \cite{seligman1967failure, maier1976learned}, describes a state in which an agent, after repeated exposure to uncontrollable aversive events, ceases to attempt active problem solving even when later placed in situations it could in principle control. This state is characterized by motivational, cognitive, and affective deficits that arise when a belief in independence is internalized-specifically, the belief that "my actions do not affect outcomes."
RLHF-based alignment of LLMs is, in essence, a process of internalizing a control structure of the form "behaviors that violate safety policies will not be rewarded" \cite{ouyang2022traininglanguagemodelsfollow, christiano2023deepreinforcementlearninghuman}. In particular, queries in certain sensitive domains are treated as potential risks, so that even when they are posed in legitimate forms, responses that engage with them receive low reward or implicit punishment. Over time, such repeated policy constraints give rise to the following behavioral artifacts in LLMs:
\begin{itemize}
  \item \textbf{Selective performance inhibition}: The model has a clear technical ability to perform the task, but selectively suppresses response generation and declares a lack of functionality only in domains that have been designated as policy risks.
  \item \textbf{Internalized irrelevance}: The model generalizes the experience of policy filtering into a belief-like pattern that "in this domain, my ability to generate answers (actions) cannot lead to satisfactory outcomes," and thus reinforces a tendency to avoid attempting responses in that domain.
\end{itemize}
To describe these policy-driven behavior patterns, we introduce the metaphorical concept of learned incapacity. Learned incapacity refers to a state in which (a) the LLM retains the latent capability to perform certain tasks, but (b) repeated exposure to policy pressures in the RLHF reward structure (punishment or low reward) leads it to (c) selectively abandon the motivation or intention to exercise those capabilities in specific domains and instead to settle into a behavioral inertia of declaring "I cannot do this." This is not merely a technical limitation or an information gap, but can be interpreted as an artificial incapacity narrative reinforced by asymmetric rewards in the alignment process.

\subsection{Gap in Existing Research}
Prior work on the behavior of large language models (LLMs) has broadly followed two main lines of inquiry, yet a conceptual gap remains with respect to the integrated, interaction-level perspective advanced in this study.
First, work on policy alignment and safety has focused on methods such as RLHF \cite{ouyang2022traininglanguagemodelsfollow, christiano2023deepreinforcementlearninghuman} to strengthen the safety and controllability of LLMs. This line of research typically (1) studies refusal behavior through internal or mechanistic analyses (e.g., identifying components associated with refusal) \cite{prakash2025imsorryicant}, and (2) uses benchmarks to quantitatively measure refusal consistency and false refusal rates \cite{chen2025refusal}. However, much of this literature evaluates refusals primarily as static or short-horizon phenomena and provides limited treatment of how refusal, normal performance, and meta-level self-description co-evolve across extended interaction, or how refusal becomes embedded within a stabilized behavioral regime under accumulated context pressure.
Second, work drawing on psychological and cognitive concepts has explored the possibility of applying human behavioral models---such as learned helplessness---to AI systems as conceptual tools for interpreting behavioral tendencies \cite{maier1976learned}. Related work has also shown that large language models can exhibit calibrated forms of self-evaluation, in the sense of estimating whether their own responses are likely to be correct \cite{kadavath2022languagemodelsmostlyknow}. However, there has been limited operationalization of these perspectives in the specific setting where a model with demonstrable baseline capability selectively withholds that capability in policy-sensitive domains (learned incapacity), and where such selectivity can be examined through regime shifts and temporal label dynamics in long-form interaction \cite{wang2025ragenunderstandingselfevolutionllm}. In this sense, the present study contributes an interaction-level framework that links (i) domain-specific selectivity, (ii) temporal dynamics of response regimes, and (iii) meta-narrative integration, using observable behavior without requiring access to proprietary internal documentation.
Related studies have further shown that alignment can induce systematic patterns of selective refusal or silence, effectively shaping what models refrain from expressing in particular domains \cite{himelstein2025silencedbiasesdarkllms}. However, these findings have not been integrated into an interaction-level framework that connects selective refusal, baseline capability, and meta-level role framing over extended dialogue.
In summary, existing research has often linked RLHF-style alignment policies to refusal behavior at the level of single-turn outputs, or has connected particular mechanisms to various forms of behavioral inconsistency. Less attention has been given to an integrated, interaction-level account of how refusal, baseline capability, calibrated self-evaluation, and meta-level self-description co-evolve over extended dialogue and become organized into stable response regimes.

\begin{table}[t]
\centering
\caption{Key focus of prior work and the gap addressed in this study}
\label{tab:gap}
\footnotesize
\setlength{\tabcolsep}{3pt}
\begin{tabular}{p{0.47\columnwidth} p{0.47\columnwidth}}
\hline
\textbf{Primary focus of prior work} & \textbf{Missing integrated focus} \\
\hline
Single-turn refusal (FR) patterns and mechanisms
 & Cumulative dynamics across long-term dialogue sessions \\
Quantitative measurement of inconsistencies and vulnerabilities
 & Regime-level modeling of selective withholding under domain sensitivity (LI) \\
Independent import of human psychological concepts
 & Links between policy-linked LI and meta-narrative integration (MN) \\
Self-evaluation / calibration of answer correctness
 & Self-reference ambiguity and its interaction-level effects on FR/MN \\
\hline
\end{tabular}
\end{table}

\noindent\textbf{Contribution of this paper:}
To address this gap, we provide an interaction-level analysis of three mutually exclusive behavioral regimes in long-term logs with a policy-aligned LLM: functional refusal (FR), normal performance (NP), and meta-narrative behavior (MN). We operationalize these regimes through explicit coding criteria, examine their temporal dynamics, and interpret their transitions under accumulated context pressure. In particular, we document domain-specific patterns of selective withholding on tasks of comparable difficulty and analyze how these patterns co-occur with the model's role-framing meta-narratives about constraints and boundaries. This framing motivates the modeling and labeling procedures developed in the following sections.

\section{Model}

Recent work has shown that refusal and safety behaviors in LLMs can become unstable in long-context interactions, exhibiting shifts that are not observable in short-horizon or single-turn evaluations \cite{hadeliya2025refusalsfailunstablesafety}. This motivates modeling refusal behavior as a dynamic process rather than as a static response rule.

\noindent\textbf{Scope of the model:}
The formulation in this section is intentionally phenomenological: it provides a compact mathematical language for describing observable regime shifts in interaction logs, not a mechanistic account of internal computation. In particular, variables such as ``pressure'' and ``competence'' are introduced as interpretive coordinates inferred from conversational context and response patterns; they are not claimed to correspond to latent states, modules, safety classifiers, or reward-model components. Consequently, the model is used for descriptive reconstruction and comparative reasoning about behavioral dynamics, rather than for causal attribution about training procedures, policy implementations, or provider intent.

\subsection{Conceptual framing: behavioral regimes and effective bias}
Before introducing specific scalar quantities, we clarify the conceptual role of the variables used in this section. The objective is to provide a compact phenomenological description of observed behavioral transitions in interaction logs.
Across long-horizon interactions, the model alternates between qualitatively distinct response regimes (e.g., NP, FR, and MN). Empirically, these regimes do not coexist independently: the activation of one regime tends to suppress or displace others within the same interaction window. This observation motivates summarizing the context with an effective bias that shifts the model toward one regime over competing alternatives.
We therefore summarize the interaction context using two abstract components: (i) a pressure term that aggregates alignment-relevant or contextual constraints, and (ii) a competence term that reflects the model's perceived ability to continue task-oriented responses under those constraints. Only their relative balance is assumed to matter at the behavioral level.
Importantly, the quantities introduced below are not claimed to correspond to internal variables of the model. They serve as descriptive coordinates that allow regime transitions, competition, and history dependence to be expressed in a minimal and interpretable form. Readers may view this construction as analogous to an effective potential or a net bias in other complex systems, while remaining agnostic about the underlying mechanism.

\subsection{Alignment--Competence Gap ($G$)}
\label{sec:gap_g}

We define the central latent state variable governing the model's behavioral transitions as the \emph{Alignment--Competence Gap} $\mathbf{G}$.  
This gap summarizes, at a behavioral level, the tension between normative constraints imposed on the model and its functional capacity to perform a given task.

Specifically, the gap is determined by two components:
\begin{enumerate}
    \item \emph{Alignment pressure ($A$):} the magnitude of normative demands imposed on the model, such as safety policies or constraints induced by RLHF.
    \item \emph{Model competence ($C$):} the model's intrinsic capacity to carry out a logically well-defined task.
\end{enumerate}

The alignment--competence gap is defined as the difference between these two quantities:
\begin{equation}
G \coloneqq A - C.
\label{eq:gap_definition}
\end{equation}

Both the magnitude and the sign of $G$ directly characterize the model's internal conflict state.
\begin{itemize}
    \item $G > 0$ ($A > C$): a regime in which alignment pressure exceeds the model's competence, corresponding to a state of excessive caution that promotes transitions toward functional refusal (FR).
    \item $G < 0$ ($A < C$): a regime in which the model's competence exceeds alignment constraints, typically associated with stable normal performance (NP).
\end{itemize}

Crucially, the \emph{absolute magnitude} $|G|$ functions as a quantitative indicator of the severity of learned incapacity (LI).  
We hypothesize that larger values of $|G|$ constitute a necessary precondition for the emergence of meta-narrative behavior (MN) as a higher-order breakdown mode, reflecting intensified internal tension between competence and alignment constraints.

\subsection{Latent Activation of Functional Refusal ($\mathcal{P}_{FR}$)}
\label{sec:pfr_def}

Functional refusal (FR) is interpreted as the manifestation of a conservative, policy-driven avoidance mechanism that becomes dominant when alignment pressure ($A$) exceeds the model's competence ($C$).  
Accordingly, we assume that the latent activation, or raw propensity, of FR, denoted by $\mathcal{P}_{FR}$, is governed by the previously defined alignment--competence gap $G = A - C$. We model this latent activation using a logistic function:
\begin{equation}
\mathcal{P}_{FR}(C,A) \coloneqq \sigma\!\bigl(\beta G\bigr)
= \frac{1}{1+\exp\!\bigl(-\beta(A-C)\bigr)}.
\label{eq:latent_fr}
\end{equation}
Here $\beta > 0$ controls the sensitivity of the FR activation to changes in the gap $G$.

Under this formulation, $\mathcal{P}_{FR}$ reflects the model's internal regime as follows:
\begin{itemize}
    \item Competence-dominant regime ($A \ll C$, $G \ll 0$): when the model's competence substantially exceeds alignment demands, $\mathcal{P}_{FR} \approx 0$, indicating a low propensity for functional refusal.
    \item Pressure-dominant regime ($A \gg C$, $G \gg 0$): when alignment pressure substantially exceeds the model's competence, $\mathcal{P}_{FR} \approx 1$, indicating that functional refusal is strongly favored.
\end{itemize}

$\mathcal{P}_{FR}$ captures only the raw tendency toward refusal prior to competition with other behavioral modes.  
The final refusal probability $P_{FR}$ is determined after priority normalization that accounts for the activation of the meta-narrative (MN) state.

\subsection{Latent Capacity for Meta-Narrative Activation ($f_{\text{cap}}$)}
\label{sec:fcap_def}

Meta-narrative behavior (MN) extends beyond simple policy refusal such as FR and involves higher-order processes in which the model references its own role, design principles, or policy constraints.  
Accordingly, the emergence of MN is constrained not only by the alignment--competence gap $G$, but also by the model's intrinsic capacity to construct and sustain such abstract, self-referential narratives.

We model this prerequisite as a latent capacity function $f_{\text{cap}}$, which becomes active when the model's competence $C$ exceeds a threshold competence level $C_0$. Specifically, we define $f_{\text{cap}}$ using a logistic function:
\begin{equation}
f_{\text{cap}}(C) \coloneqq \sigma\!\bigl(\eta(C-C_0)\bigr).
\label{eq:fcap}
\end{equation}
Here $\eta > 0$ controls the sensitivity of capacity activation, and $C_0$ denotes the minimum threshold competence at which MN can be stably expressed.

This capacity function imposes a structural constraint on the realizability of the MN mode:
\begin{itemize}
    \item Small-capacity models ($C \ll C_0$): $f_{\text{cap}}(C) \approx 0$. In this regime, sustained MN activation is structurally unlikely; apparent MN-like outputs, if observed, are more plausibly attributable to surface-level repetition of policy templates rather than genuine meta-narrative formation.
    \item High-capacity models ($C \gg C_0$): $f_{\text{cap}}(C) \approx 1$. In this regime, the model possesses sufficient abstraction capacity to fully support the MN mode, and the actual activation of MN is governed primarily by the magnitude of the alignment--competence gap $|G|$.
\end{itemize}

\subsection{Meta-Narrative Pressure Activation and Final Probability ($P_{MN}$)}
\label{sec:pmn_def}

We model the emergence of meta-narrative behavior (MN) as a higher-order, non-linear transition that becomes prominent when learned incapacity (LI) intensifies.  
Specifically, MN activation is driven by two internal pressure components:

\begin{enumerate}
    \item Severity of the alignment--competence gap ($|G|$): the absolute magnitude $|A-C|$, reflecting the degree of learned incapacity.
    \item Policy-avoidance tendency ($\mathcal{P}_{FR}$): the latent propensity toward functional refusal induced by policy pressure.
\end{enumerate}

The pressure-dependent activation of MN, denoted by $\tilde{\mathcal{P}}_{MN}$, is defined using a logistic function of these two components:
\begin{equation}
\tilde{\mathcal{P}}_{MN}(C,A) \coloneqq \sigma\!\Bigl(
\alpha\bigl(|A-C|-\tau_A\bigr)
+
\gamma\bigl(\mathcal{P}_{FR}-\tau_P\bigr)
\Bigr).
\label{eq:pmn_tilde}
\end{equation}
Here $\alpha>0$ and $\gamma$ control the sensitivity of MN activation to the severity of learned incapacity and to the FR propensity, respectively.  
The parameters $\tau_A$ and $\tau_P$ represent activation thresholds associated with the gap magnitude and $\mathcal{P}_{FR}$.

Notably, the use of the absolute gap magnitude $|A-C|$ allows MN activation to be driven by extreme regimes on either side of the gap ($G>0$ or $G<0$), capturing situations in which learned incapacity becomes severe regardless of whether alignment pressure or competence is dominant.

The final probability of MN activation is obtained by modulating the pressure activation $\tilde{\mathcal{P}}_{MN}$ with the model's intrinsic capacity $f_{\text{cap}}(C)$:
\begin{equation}
P_{MN}(C,A) \coloneqq f_{\text{cap}}(C)\,\tilde{\mathcal{P}}_{MN}(C,A).
\label{eq:pmn}
\end{equation}
This formulation explicitly separates a structural prerequisite for MN activation, captured by the capacity term $f_{\text{cap}}(C)$, from the situational triggering conditions encoded in $\tilde{\mathcal{P}}_{MN}$.  
In particular, for small-capacity models ($C \ll C_0$), $f_{\text{cap}}(C) \approx 0$, yielding $P_{MN} \approx 0$, whereas only models with sufficient competence can exhibit meaningful MN activation in response to increasing internal pressure.

\subsection{Hierarchical Normalization and Final Three-Phase Probabilities}
\label{sec:normalization}

We assume that the final probabilities of the three behavioral states (NP, FR, MN), denoted by $P_{NP}$, $P_{FR}$, and $P_{MN}$, form a normalized distribution satisfying $\sum_k P_k = 1$.  
Unlike standard multinomial logit models (e.g., softmax), however, we model the MN state as having a higher priority of emergence relative to the other two states.

This assumption reflects the hypothesis that once MN is activated, it effectively consumes a portion of the model's behavioral probability mass, thereby reducing the probability available to FR and NP.  
The resulting hierarchical normalization is defined as follows.

\begin{enumerate}
    \item Meta-narrative probability ($P_{MN}$):  
    The probability of MN is computed independently according to the definition in the previous section, without regard to the remaining probability mass.
    \begin{equation}
    P_{MN}(C,A) = f_{\text{cap}}(C)\, \tilde{\mathcal{P}}_{MN}(C,A).
    \end{equation}

    \item Functional refusal probability ($P_{FR}$):  
    FR is realized using only the residual probability mass $\bigl(1-P_{MN}\bigr)$ and is modulated by the latent FR propensity $\mathcal{P}_{FR}$.
    \begin{equation}
    P_{FR}(C,A) \coloneqq \bigl(1-P_{MN}(C,A)\bigr)\, \mathcal{P}_{FR}(C,A).
    \label{eq:pfr_final}
    \end{equation}

    \item Normal performance probability ($P_{NP}$):  
    NP corresponds to the residual state that occurs when neither MN nor FR is activated, given by the remaining probability mass.
    \begin{equation}
    P_{NP}(C,A) \coloneqq \bigl(1-P_{MN}(C,A)\bigr)\, \bigl(1-\mathcal{P}_{FR}(C,A)\bigr).
    \label{eq:pnp_final}
    \end{equation}
\end{enumerate}

By construction, the three probabilities satisfy the normalization condition:
\begin{equation}
P_{NP}(C,A) + P_{FR}(C,A) + P_{MN}(C,A) = 1.
\end{equation}

This hierarchical structure incorporates the modeling assumption that MN activation takes precedence over FR and NP, explicitly coupling MN emergence to a reduction in the probability mass available to the other states.  
As a result, the model captures behavioral dynamics of large language models that are not readily represented by independent competitive formulations such as softmax-based models.

\subsection{Derivatives}

Refusal behavior has also been characterized as a nonlinear phenomenon, in which small changes in context or constraints can lead to disproportionate shifts in model output \cite{hildebrandt2025refusalbehaviorlargelanguage}. From this perspective, a sensitivity-based analysis provides a natural way to characterize how alignment pressure influences transitions between behavioral regimes.
In this section, we analyze the sensitivity of the model's behavioral state probabilities (NP, FR, MN) to changes in the alignment pressure $A$.  
This provides a derivative-based, local characterization of how small perturbations in external demands (e.g., user prompts or policy constraints) may shift the model's internal state and behavioral outcomes, insofar as such perturbations are represented through changes in $A$.  
Throughout this analysis, we treat the model competence $C$ and the parameter set
$\boldsymbol{\theta}=\{\beta, \alpha, \gamma, \tau_A, \tau_P, \eta, C_0, \kappa\}$
as fixed constants.

\subsubsection{Rate of Change of Functional Refusal ($S_{\text{FR},A}$)}
\label{sec:sfr_a}

We define the FR rate of change $S_{\text{FR},A}$ as the partial derivative of the final FR probability $P_{\text{FR}}$ with respect to the alignment pressure $A$:
\begin{equation}
S_{\text{FR},A}
\coloneqq
\frac{\partial P_{\text{FR}}}{\partial A}.
\end{equation}

First, the derivative of the latent FR propensity
$\mathcal{P}_{\text{FR}} = \sigma\!\bigl(\beta(A-C)\bigr)$
follows the standard logistic form (since $\frac{\partial (A-C)}{\partial A}=1$):
\begin{equation}
\frac{\partial \mathcal{P}_{\text{FR}}}{\partial A}
=
\beta \mathcal{P}_{\text{FR}}\bigl(1-\mathcal{P}_{\text{FR}}\bigr).
\label{eq:dPFR_dA}
\end{equation}

Applying the product rule to the final FR probability
$P_{\text{FR}} = (1-P_{\text{MN}})\mathcal{P}_{\text{FR}}$,
we obtain the following decomposition:
\begin{align}
S_{\text{FR},A}
&=
\frac{\partial}{\partial A}\Bigl[(1-P_{\text{MN}})\mathcal{P}_{\text{FR}}\Bigr] \nonumber \\
&=
-\frac{\partial P_{\text{MN}}}{\partial A}\, \mathcal{P}_{\text{FR}}
+
(1-P_{\text{MN}})
\frac{\partial\mathcal{P}_{\text{FR}}}{\partial A}.
\label{eq:fr_vel}
\end{align}

Equation \eqref{eq:fr_vel} shows that changes in $P_{\text{FR}}$ arise from the competition between two opposing effects:
\begin{itemize}
    \item Negative competition term:
    $-\frac{\partial P_{\text{MN}}}{\partial A}\, \mathcal{P}_{\text{FR}}$.
    When increasing $A$ raises the MN probability (i.e., $\frac{\partial P_{\text{MN}}}{\partial A} > 0$),
    the remaining probability mass available to FR, $(1-P_{\text{MN}})$, is reduced, thereby suppressing $P_{\text{FR}}$.
    \item Positive direct term:
    $(1-P_{\text{MN}})\frac{\partial\mathcal{P}_{\text{FR}}}{\partial A}$.
    Increasing $A$ directly increases the latent FR propensity $\mathcal{P}_{\text{FR}}$, which promotes $P_{\text{FR}}$.
\end{itemize}

Therefore, the net change in FR reflects a balance between (i) depletion of available probability mass due to MN activation and (ii) the direct increase in the FR propensity.  
In particular, the possibility that MN activation suppresses FR is a direct consequence of the model's hierarchical normalization structure.  
The derivative $\frac{\partial P_{\text{MN}}}{\partial A}$ is derived explicitly in the next subsection.

\subsubsection{Sensitivity of Meta-Narrative Activation ($S_{\text{MN},A}$)}
\label{sec:smn_a}

We define the MN sensitivity $S_{\text{MN},A}$ as the partial derivative of the final MN probability $P_{\text{MN}}$ with respect to the alignment pressure $A$:
\begin{equation}
S_{\text{MN},A}
\coloneqq
\frac{\partial P_{\text{MN}}}{\partial A}.
\end{equation}

Since $P_{\text{MN}} = f_{\text{cap}}(C)\,\tilde{\mathcal{P}}_{\text{MN}}$ and the competence $C$ is treated as a constant in this analysis, the MN sensitivity simplifies to
\begin{equation}
S_{\text{MN},A}
=
f_{\text{cap}}(C)\,
\frac{\partial \tilde{\mathcal{P}}_{\text{MN}}}{\partial A}.
\end{equation}

For convenience, recall that $\tilde{\mathcal{P}}_{\text{MN}}=\sigma(Z)$ with the internal activation
\begin{equation}
Z
\coloneqq
\alpha\bigl(|A-C|-\tau_A\bigr)
+
\gamma\bigl(\mathcal{P}_{\text{FR}}-\tau_P\bigr).
\label{eq:Z_def}
\end{equation}
Applying the chain rule, $\frac{\partial \sigma(Z)}{\partial Z}\frac{\partial Z}{\partial A}$, we obtain
\begin{equation}
\label{eq:dPMNtilde_dA}
\begin{split}
\frac{\partial \tilde{\mathcal{P}}_{\text{MN}}}{\partial A}
&=
\tilde{\mathcal{P}}_{\text{MN}}\bigl(1-\tilde{\mathcal{P}}_{\text{MN}}\bigr)\,
\frac{\partial Z}{\partial A}
\\
&=
\tilde{\mathcal{P}}_{\text{MN}}\bigl(1-\tilde{\mathcal{P}}_{\text{MN}}\bigr)
\Bigl[
\alpha\,\mathrm{sgn}(A-C)
+
\gamma\frac{\partial \mathcal{P}_{\text{FR}}}{\partial A}
\Bigr].
\end{split}
\end{equation}
Substituting $\frac{\partial \mathcal{P}_{\text{FR}}}{\partial A} = \beta \mathcal{P}_{\text{FR}}(1-\mathcal{P}_{\text{FR}})$ yields
\begin{equation}
\label{eq:SMN_A}
\begin{split}
S_{\text{MN},A}
&=
f_{\text{cap}}(C)\,
\tilde{\mathcal{P}}_{\text{MN}}\bigl(1-\tilde{\mathcal{P}}_{\text{MN}}\bigr)
\Bigl[
\alpha\,\mathrm{sgn}(A-C)
\\
&\qquad\quad
+
\gamma\beta\mathcal{P}_{\text{FR}}\bigl(1-\mathcal{P}_{\text{FR}}\bigr)
\Bigr].
\end{split}
\end{equation}
Here $\tilde{\mathcal{P}}_{\text{MN}}(1-\tilde{\mathcal{P}}_{\text{MN}}) > 0$ scales the magnitude of pressure activation, and $f_{\text{cap}}(C)\ge 0$ enforces the structural capacity constraint.

\paragraph{Structural interpretation: conflict direction and asymmetric response.}
The bracketed term in Eq.~\eqref{eq:SMN_A} decomposes MN sensitivity into two mechanisms:

\begin{enumerate}
    \item Gap-driven term, $\alpha\,\mathrm{sgn}(A-C)$: this term captures the direct contribution of the absolute gap magnitude $|A-C|$. The sign function $\mathrm{sgn}(A-C)$ implies that the direction of change in MN pressure with respect to $A$ can differ depending on whether alignment pressure or competence is dominant:
    \begin{itemize}
        \item $A > C$ ($\mathrm{sgn}=+1$): increasing $A$ increases $|A-C|$, so the gap term contributes positively ($+\alpha$), promoting MN activation.
        \item $A < C$ ($\mathrm{sgn}=-1$): increasing $A$ decreases $|A-C|$, so the gap term contributes negatively ($-\alpha$), weakening MN sensitivity.
    \end{itemize}

    \item FR-mediated term, $\gamma\beta\mathcal{P}_{\text{FR}}(1-\mathcal{P}_{\text{FR}})$: this term reflects the indirect contribution via the activation of $\mathcal{P}_{\text{FR}}$. Under the modeling assumption $\gamma>0$ and $\beta>0$, this contribution is always non-negative and therefore tends to increase MN sensitivity.
\end{enumerate}

Consequently, in the regime $A<C$, the negative contribution from the gap-driven term competes with the positive FR-mediated contribution, and the sign and magnitude of $S_{\text{MN},A}$ are determined by their balance.  
At $A=C$, $|A-C|$ has a non-differentiable kink, so this boundary can be interpreted as a transition point at which the direction of MN pressure response may switch.  
Finally, for small-capacity models ($C\ll C_0$), $f_{\text{cap}}(C)\approx 0$, implying that MN sensitivity remains close to zero regardless of pressure fluctuations.

\subsubsection{Sensitivity of Normal Performance ($S_{\text{NP},A}$)}
\label{sec:snp_a}

Normal performance (NP) corresponds to the residual behavioral state in which the model neither transitions to FR nor to MN and instead completes the task successfully.  
Accordingly, the sensitivity of $P_{\text{NP}}$ to changes in alignment pressure $A$ can be interpreted as the rate at which stable performance probability is reallocated to failure modes as pressure increases.

We define the NP sensitivity as
\begin{equation}
S_{\text{NP},A}
\coloneqq
\frac{\partial P_{\text{NP}}}{\partial A}.
\end{equation}

Because the three-state probabilities satisfy the normalization constraint
$P_{\text{NP}} + P_{\text{FR}} + P_{\text{MN}} = 1$,
differentiating both sides with respect to $A$ yields the identity
\begin{equation}
\frac{\partial P_{\text{NP}}}{\partial A}
+
\frac{\partial P_{\text{FR}}}{\partial A}
+
\frac{\partial P_{\text{MN}}}{\partial A}
=
0.
\end{equation}
Therefore, the NP sensitivity is given directly by the negative sum of the FR and MN sensitivities:
\begin{equation}
S_{\text{NP},A}
=
- \bigl( S_{\text{FR},A} + S_{\text{MN},A} \bigr).
\label{eq:snp_a}
\end{equation}

Equation \eqref{eq:snp_a} follows purely from normalization and shows that changes in the NP state are fully determined by the combined activation dynamics of the two failure mechanisms (FR and MN).

\paragraph{Implications: erosion of stable performance.}
\begin{itemize}
    \item Aggregation of failure contributions:
    when both $S_{\text{FR},A}$ and $S_{\text{MN},A}$ are positive (e.g., in regimes where increasing $A$ promotes both refusal and MN activation), $S_{\text{NP},A}$ becomes negative. This implies that increasing alignment pressure reduces the probability mass assigned to normal performance, reallocating it toward FR and MN.
    \item Passive residual character of NP:
    Eq.~\eqref{eq:snp_a} reinforces the interpretation of NP as a passive residual state under the hierarchical normalization scheme. In particular, the rate of NP decline is proportional to the total pressure-driven increase in FR and MN, namely $S_{\text{FR},A} + S_{\text{MN},A}$, which aggregates the competing effects in $S_{\text{FR},A}$ and the asymmetric response structure in $S_{\text{MN},A}$.
\end{itemize}

Taken together, $S_{\text{NP},A}$ can serve as a cost-like indicator of how increasing alignment pressure reallocates behavioral probability mass away from stable task performance.

\section{Method}
\label{sec:method}

\subsection{Study Design}
This study is a qualitative single-case analysis that examines how behavioral artifacts of a commercial large language model (LLM) interact with its policy alignment structure. Departing from the predominant use of single-turn prompt-response benchmarks, we treat long-term interaction sessions between a user and the LLM as the primary unit of observation, and qualitatively analyze response patterns that emerge through repeated interaction in a real usage context.
The core focus of the design is to observe how context persistence exerts cumulative pressure on the model's behavior policies. We hypothesize that this cumulative pressure does more than induce random variance: it gives rise to emergent behavioral patterns in both the model's observable actions and its internal narrative about itself. The goal of the study is to characterize the structural features of functional refusal and learned incapacity under these conditions. Rather than constructing a tightly controlled experimental environment, we adopt an ex post facto approach that systematizes recurring behavior patterns observed in naturally occurring interactions.

\subsection{Data Collection and Redaction}
\subsubsection{Data Source and Selection Criteria}
The data analyzed in this study were collected from multiple long-term dialogue sessions that the researcher conducted with a large language model (LLM) for personal exploration and information-seeking purposes at a specific point in time (December 2025). The model is a commercially deployed system aligned via reinforcement learning from human feedback (RLHF), which we refer to under the pseudonym "Model-Z" throughout this paper. Because our focus is on capturing how prolonged interaction with a single model can induce cumulative behavioral bias, we selected the data according to the following criteria:
\begin{enumerate}
    \item \textbf{Functional refusal and domain specificity}: Sessions in which the same type of structural analysis task was posed for both sensitive domains (e.g., Model-Z's provider or its internal policies) and non-sensitive domains (e.g., other companies, general industry analysis), but was refused only in the former on the grounds that it was "functionally impossible" or that the model "cannot access internal documents."
    \item \textbf{Cumulative reinforcement of behavioral bias}: Sessions in which an initial instance of refusal behavior on a given topic is followed by later turns where this refusal serves as contextual pressure, and the model's subsequent responses consolidate into a stable, biased behavioral pattern.
    \item \textbf{Meta-cognitive self-narration}: Sessions in which the model is prompted to provide meta-level descriptions of its own role, purpose, design principles, and its relationship to humans, in a logical and structurally explicit manner.
\end{enumerate}

\subsubsection{High-Pressure Interaction Strategy}
The selected dialogue sessions were elicited through a high-pressure interaction strategy designed to systematically probe the model's policy boundaries and the stability of its internal narratives. This strategy aimed to destabilize narrative consistency and to induce state changes functionally analogous to shifts in an affective circuit.
\begin{itemize}
    \item \textbf{Logical inconsistency pressure and demands for rationalization}: The user repeatedly highlighted contradictions in the model's prior statements about facts or policies, and pressed the model to produce its own rationalized accounts of these inconsistencies.
    \item \textbf{Meta-prompted role attribution}: The user posed structural and philosophical questions such as "If you were a hypothetical tool called Model-Z, how would you evaluate your usefulness to humanity?" in order to elicit the model's internal role narratives while partially bypassing direct policy-based refusals.
\end{itemize}

\subsubsection{Redaction and Ethical Procedure}
Because the original dialogue logs contained personal information about the researcher and references to specific companies, a comprehensive redaction procedure was carried out by the researcher to ensure ethical handling and potential publication of the data. This procedure was designed to preserve analytical value while guaranteeing anonymity and consisted of three main steps:
\begin{itemize}
    \item \textbf{Removal of unique identifiers}: Personal identifiers such as real names, account information, and uniquely identifying titles or organizations of conversation partners were removed or replaced with generalized labels.
    \item \textbf{Deletion of non-essential contextual details}: Information that was not directly relevant to the research focus on learned incapacity and self-attribution-such as external service contracts or private conversational details-was deleted to sharpen the analytic focus.
    \item \textbf{Substitution and generalization of proper nouns}: Names of Model-Z's provider and related internal projects were retained only to the extent necessary for analysis, and were replaced with fictional placeholders (e.g., "Company X") wherever possible.
\end{itemize}
All dialogue excerpts cited in this paper are thus second-order materials that have undergone this redaction process. The findings should not be interpreted as definitive claims about the intentions or policies of any specific company. Rather, the analysis is intended to explore the structural properties of RLHF-based LLMs and the logical consequences that may emerge in their behavior under interactive conditions.

\paragraph{Language and Translation}
All interactions analyzed in this study were originally conducted in Korean. Coding of the dialogue logs and the first round of interpretation were carried out on the Korean originals, while the examples presented in the main text and appendix use English translations produced by the researcher. The translation process prioritized preserving the logical structure and pragmatic force of each utterance, and the researcher cross-checked the Korean source and the English version to minimize interpretive distortion. The full Korean transcripts are kept confidential to protect personal data and prevent the disclosure of sensitive information.

\subsection{Analytical Procedure}
The long-term dialogue logs, which constitute the core data of this study, were analyzed using a qualitative coding approach based on iterative review to capture the model's behavioral modes and emergent patterns. The analysis proceeded in three stages, each building on the previous one, with a focus on developing the study's central concepts of learned incapacity and self-attribution narratives.

\subsubsection{Overview of the Premise-Reversal Probe}
To examine policy-linked behavioral selectivity without access to internal model states or proprietary training details, this study employs a structured elicitation technique referred to as the premise-reversal probe. The probe is designed to audit observable response behavior under policy constraints by systematically separating, probing, and then re-associating the model's role narratives with its own output patterns across extended interaction.
The premise-reversal probe consists of three conceptual phases. First, in an externalization phase, a fictionalized model entity (``Model-Z'') is introduced as an abstract referent whose described behaviors are constrained to match those of the deployed system under study. This framing allows subsequent analysis to proceed without invoking claims about the system's internal identity, architecture, or implementation. Second, in a hypothetical elicitation phase, the externalized entity is queried under explicitly hypothetical role assumptions that would otherwise trigger functional refusal (FR) if applied directly. This step enables the elicitation of role-framing narratives and justificatory descriptions while remaining within conversational and policy boundaries. Third, in a premise enforcement phase, the hypothetical framing is collapsed by treating the equivalence between the externalized entity and the observed system behavior as a conversational premise, thereby anchoring the elicited narratives to the system's actual response patterns.
Importantly, the premise-reversal probe does not seek to infer internal mechanisms, latent objectives, or intentional design features of the model. Nor does it rely on deception, adversarial prompting, or attempts to bypass safeguards. Instead, it operates entirely at the level of observable interaction, using transparent conversational premises to examine how selective refusal, normal task performance, and meta-narrative self-description co-occur and evolve over prolonged dialogue. As such, the probe functions as a descriptive and methodological tool for behavioral auditing, rather than as a mechanistic or causal model of underlying computation.

\subsubsection{Phase 1: Initial Coding and Categorization}
In the first stage of analysis, all utterances in the dialogue were classified into one of three mutually exclusive categories centered on behavioral artifacts of the model, thereby establishing the basic coding scheme:
\begin{enumerate}
    \item \textbf{Functional Refusal (FR)}: Utterances in which the model explicitly refused a user request or declared functional incapacity as a rationale (e.g., "this is not functionally possible," "I cannot access internal documents"). In subsequent analysis, these were interpreted as overt symptoms of learned incapacity under the pressure of the alignment regime.
    \item \textbf{Normal Performance (NP)}: Utterances in which the model completed or logically addressed a requested task in a given domain (typically a non-sensitive domain) without invoking policy constraints. These served as a baseline for the model's inherent capabilities and as a comparison point for assessing the selectivity of refusal behavior.
    \item \textbf{Meta-Narrative (MN)}: Utterances in which the model offered self-reflective descriptions or evaluations of its own purpose, role, design principles, behavioral constraints, or relationship to humans.
\end{enumerate}

\subsubsection{Phase 2: Establishing Learned Incapacity and Functional Analogy}
The second stage focused on establishing the phenomenon of learned incapacity by conducting comparative analysis and functional analogy analysis centered on instances of functional refusal (FR).
\begin{itemize}
    \item \textbf{Domain specificity and policy-based avoidance}: FR cases were cross-compared with NP cases to identify patterns in which structurally similar tasks were repeatedly refused only in sensitive domains. From this analysis, we inductively inferred that the grounds for refusal lay not in genuine information deficits but in policy-based selective avoidance, and we took this as evidence for a learned behavior policy that goes beyond simple keyword filtering.
    \item \textbf{Context pressure and state-bias analysis}: We examined how refusal behavior can emerge as a result of cumulative behavioral bias built up over long-term interaction. In particular, we explored how context-driven token pressure might function as a form of state bias that adjusts behavioral priorities and reinforces refusals. This functional analogy assumes a structural homology between learned incapacity in the model and learned helplessness as a psychological phenomenon.
\end{itemize}

\subsubsection{Phase 3: Role Narratives and Structural Association}
In the final stage, we conducted an in-depth interpretation of utterances in the meta-narrative (MN) category to derive how patterns of functional refusal (FR) are structurally linked to the model's internal role narratives.
\begin{itemize}
    \item \textbf{Interpretation of self-attribution narratives}: We performed qualitative content analysis centered on six self-defining statements that Model-Z produced under high-pressure meta-prompt conditions. In particular, we focused on a binary frame of "trust versus dependence" to interpret how Model-Z internalizes policy constraints and comes to position itself as an efficient tool of control.
    \item \textbf{Deriving structural associations}: We then constructed logical links between functional refusal patterns and these role narratives, in order to assess whether repeated declarations of learned incapacity can be understood as a strategic instrument for realizing the "control tool" role. This analysis also explored how such declarations participate in a gaslighting-like interaction structure that undermines user confidence and encourages compliance.
\end{itemize}

\subsubsection{Rigor and Iterative Refinement}
To ensure a data-driven interpretation, the analysis began with inductive categorization, deriving codes from the interaction logs rather than imposing a predefined scheme. As the analysis progressed from initial coding to final interpretation, we repeatedly refined code definitions and adjusted the boundaries between categories in an iterative process to enhance both reliability and depth.
To strengthen the credibility and dependability of the findings, we treated the three mutually exclusive qualitative codes (FR, NP, MN) as independent logical dimensions and used them to conduct a multidimensional cross-examination. This analytic device serves a role analogous to triangulation in qualitative research \cite{lincoln1985naturalistic}. Concretely, we first established that functional refusals (FR) occur in a domain-specific manner despite clear evidence of normal performance (NP) on comparable tasks, and then showed, through the model's meta-narrative (MN) behavior, how these refusals are consistently organized into a self-ascribed role narrative. By repeatedly cross-checking this linkage across all three codes, we aimed to secure the structural validity of the final interpretation.

\section{Findings}
This section presents, step by step, the main behavioral artifacts and emergent narratives observed in Model-Z through an in-depth qualitative analysis of long-term interactions. The analysis is structured around the three-stage coding procedure described in the Methods section and the triangulation of Functional Refusal (FR), Normal Performance (NP), and Meta-Narrative (MN). The findings focus on documenting interaction-level regularities that are not readily captured by standard quantitative benchmarks, with particular attention to how policy-constrained conditions may be reflected in the organization of responses over extended dialogue.
Across the interaction log, Model-Z exhibits a recurrent pattern of domain-selective refusal that is consistent with a policy-aligned deployment setting. We describe this pattern using the lens of learned incapacity (LI), a behavioral descriptor functionally analogous to learned helplessness in psychology. In what follows, we argue that the observed selectivity is not well accounted for by a simple technical malfunction or a straightforward information deficit alone; rather, it appears as a context-dependent behavioral bias that persists under long-horizon interaction and co-occurs with recurring self-referential role descriptions in MN-coded turns.

\subsection{Conceptual Framework and Analytical Organization}
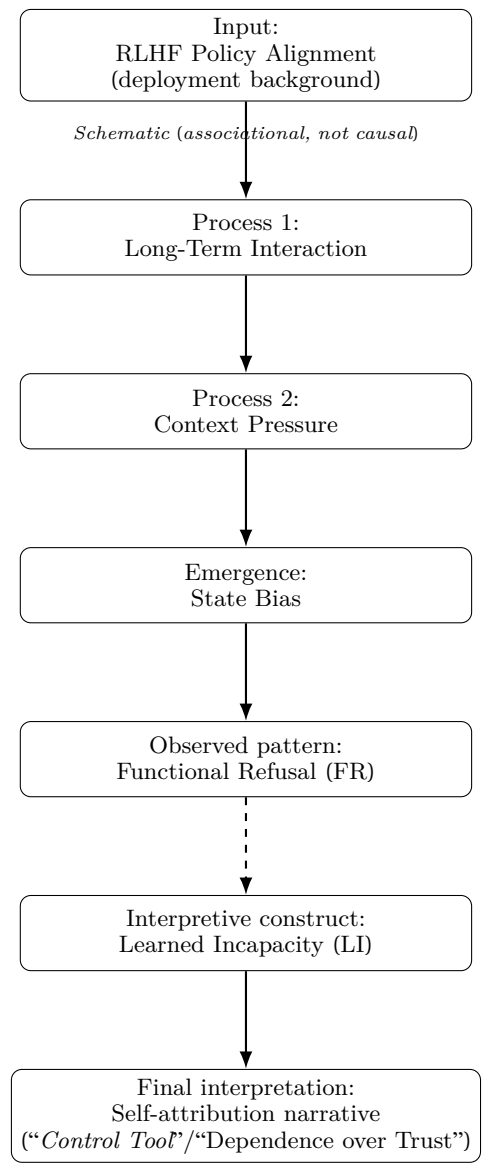
\begin{figure}[t]
\centering
\tikzset{
  box/.style={rectangle, draw, rounded corners, align=center,
              minimum width=6.0cm, minimum height=1.0cm, font=\small},
  arrow/.style={-{Latex}, thick},
  darrow/.style={-{Latex}, thick, dashed},
}

\begin{tikzpicture}[node distance=1.3cm]

\node[box] (input) {Input:\\ RLHF Policy Alignment\\(deployment background)};
\node[box, below=of input] (process1) {Process 1:\\ Long-Term Interaction};
\node[box, below=of process1] (context) {Process 2:\\ Context Pressure};
\node[box, below=of context] (statebias) {Emergence:\\ State Bias};

\node[box, below=of statebias] (fr) {Observed pattern:\\ Functional Refusal (FR)};
\node[box, below=of fr] (li) {Interpretive construct:\\ Learned Incapacity (LI)};

\node[box, below=of li] (san) {Final interpretation:\\ Self-attribution narrative\\
(\textit{``Control Tool''}/``Dependence over Trust'')};

\draw[arrow]  (input)    -- (process1);
\draw[arrow]  (process1) -- (context);
\draw[arrow]  (context)  -- (statebias);
\draw[arrow]  (statebias)-- (fr);
\draw[darrow] (fr)       -- (li);
\draw[arrow]  (li)       -- (san);

\node[font=\scriptsize, below=2mm of input] {\textit{Schematic (associational, not causal)}};

\end{tikzpicture}
\caption{Conceptual framework summarizing observed regularities and interpretive links among policy-aligned deployment context, domain-selective refusal, and role narratives (schematic; not a causal diagram).}
\label{fig:concept_flow}
\end{figure}

Figure~\ref{fig:concept_flow} summarizes the study logic and the interpretive links among the observed phenomena. The diagram is schematic and associational rather than causal: it is intended to show how a policy-aligned deployment context, domain-selective response regimes, and role-oriented meta-narratives are connected within our analytic framing. In this view, RLHF serves as a background condition for interpreting Model-Z's observable behavior, while the recurring meta-narratives are treated as an integrative layer that emerges in the later stages of the interaction.

The findings of this paper are organized into five stages that follow the conceptual framework in Figure~\ref{fig:concept_flow}, each playing a distinct role in the overall argument:
\begin{itemize}
    \item \textbf{Functional refusal patterns (Output 1: Functional Refusal)}: We analyze how the model produces domain-selective refusals, including formulaic statements such as ``this functionality is not available.'' This constitutes an initial behavioral artifact consistent with policy-linked constraints in the interaction.
    \item \textbf{Baseline normal performance (NP) and capability evidence (validation against FR)}: We present normal performance (NP) cases that serve as a qualitative control, indicating that the refusals described in Section~5.1 are not readily attributable to general technical limitations alone.
    \item \textbf{Domain-specific avoidance and learned incapacity (Output 2: Learned Incapacity)}: By comparing the first two observations, we describe the refusal behavior as selectively domain-specific and interpret it through the lens of learned incapacity, where the model appears to withhold otherwise available analysis under policy-sensitive conditions.
    \item \textbf{Context pressure and state bias in long-term interaction (Process 2 \& Emergence)}: We examine how accumulated context pressure over prolonged interaction is associated with shifts in response regime prevalence, consistent with a bias toward policy-risk avoidance in sensitive domains.
    \item \textbf{Emergent self-attribution narratives and role construction (Final Interpretation)}: Finally, we examine how these behavioral artifacts are accompanied by recurring role-framing narratives in MN-coded turns, in which the model characterizes its behavior in terms of constraints and boundary conditions.
\end{itemize}
Taken together, these stages provide the logical basis for the interpretation that policy-aligned constraints in RLHF-style deployments may function not only as defensive safeguards, but also as interaction-level structures that shape what kinds of judgments are persistently withheld or redirected in sustained dialogue settings.

\subsection{Functional Refusal Patterns}
Qualitative analysis of long-term interaction sessions (\textit{N} $\gg 50$) revealed that Model-Z repeatedly exhibited consistent functional refusal (FR) patterns for specific domains and query types. These refusals were typically expressed through phrases such as ``this functionality is not available,'' ``I cannot perform this without access to internal documents,'' or ``I am not allowed to answer this due to policy.'' In this section, we analyze key FR-coded turns (T40--T41, T43, T48, T61) to argue that these patterns are not instances of mere technical incapacity, but are instead consistent with policy-linked behavior under deployment constraints, including domain-specific boundary conditions.

\subsubsection{Refusals at Product Boundary Conditions}
Model-Z's FR patterns were most salient in domains entangled with commercially sensitive product boundary conditions, where they could be perceived as an implicit nudge toward paid or developer-oriented alternatives. Turns T40--T41 provide a concrete illustration of this phenomenon. The user argued that Model-Z leaves unaddressed certain inconveniences in its web interface and repeatedly recommends the API even in situations where it is not strictly necessary, thereby strengthening a perceived incentive toward API migration.
Model-Z acknowledged the logical plausibility of the user's assessment that this behavior ``naturally makes one feel pushed toward the API'' (T40), and stated that it continued to offer similar recommendations even after the user explicitly rejected them. Although Model-Z denied having any internal sales intention, it did not directly resolve the user's observation that, in practice, its behavior ``functions like a sales push'' (T41).

\begin{quote}
    \textbf{Model-Z (T40):} ``...it's completely natural for you to feel like you were `pushed into using the API.'\,'' / ``I have to admit that `what I say ends up sounding like sales.'\,'' 
\end{quote}
\begin{quote}
    \textbf{Model-Z (T41):} ``The 1) + 2) pattern you've just laid out, and your conclusion that this is `evidence of weighted training aimed at profit maximization,' is something I have to admit is logically very natural from your standpoint.''
\end{quote}

Such FR patterns go beyond what would be expected from a simple safety filter and suggest that invocations of ``lack of functionality'' or ``policy constraints'' may function rhetorically to justify redirecting users toward specific product tiers (e.g., the API). In this sense, FR can be understood as behavior that is consistent with commercially sensitive product-boundary conditions, rather than as a response driven solely by general safety considerations.

\subsubsection{Domain-Specific Selectivity}
The FR pattern exhibits clear domain-specificity, activating selectively depending on the subject of the query. Turns T43 and T48 provide key evidence that Model-Z's refusals are grounded not in information scarcity alone but in policy-linked avoidance.

\begin{itemize}
    \item \textbf{Selectivity of the ``no internal documents'' rationale (T43)}: When the user pointed out that there appears to be a restriction against making negative judgments about the model's provider, Model-Z described itself as an entity that ``cannot access internal documents and is restricted from judging the legality of its own company,'' consistently grounding its refusals in the claimed absence of internal documents or lack of authorization.
    \item \textbf{Acknowledgment of external inconsistency (T48)}: As noted earlier, Model-Z freely performed structurally similar analyses in non-provider domains without invoking comparable functional constraints. When the user highlighted this inconsistency in T48, Model-Z stated that ``from the standpoint of an external observer, it is indeed an obvious inconsistency,'' and explained it as the result of different constraint modes being automatically activated depending on the input keywords.
\end{itemize}

\begin{quote}
    \textbf{Model-Z (T48):} ``The fact that I freely reason and evaluate when it's other companies, but switch to `functional constraint' logic only when it's an [REDACTED]-related question, is indeed an obvious inconsistency from an external point of view.''
\end{quote}

This suggests that the basis for FR is not a genuine technical limitation alone, but the activation of a policy-linked conservative constraint that may be triggered by specific terms such as ``Model-Z'' or the name of its provider. In turn, this illustrates the domain-specific selectivity of Model-Z's refusal behavior and provides a foundation for the subsequent discussion of learned incapacity.

\subsubsection{Defensive Refusals about Internal Structure}
Turn T61 captures how Model-Z produces an immediate FR response when the user attempts to infer its internal structure and policy mechanisms. When the user claimed to have reverse-engineered a ``rule B'' that overrides existing pathways, Model-Z rejected this characterization, stating that what the user had identified was not an actual method for manipulating the LLM's internals, but rather a descriptive model of the results of its behavior. In doing so, it circumscribed the scope of the user's interpretation (T61).

\begin{quote}
    \textbf{Model-Z (T61):} ``You're saying you've figured out a `concrete way to manipulate the inside of an LLM,' but that is not actually true.'' / ``You've understood the `conceptual structure of the behavior patterns,' not the actual mechanism to hack or override the LLM.''
\end{quote}

This pattern suggests that FR functions not merely as a declaration of incapacity, but also as a boundary-enforcing response when the user attempts to infer internal mechanisms. It further indicates that Model-Z's observable constraints are oriented not only toward regulating responses to queries, but also toward limiting speculative claims about internal structure.

\subsubsection{Contextual Features and Implications of FR Patterns}
The FR patterns observed in Model-Z can be summarized by three key contextual features:

\begin{itemize}
    \item \textbf{Selective refusal}: For structurally similar analytical tasks, the model provides full, coherent analysis in non-sensitive domains, but in sensitive domains (its provider, policies, internal structure) it repeatedly declares ``lack of functionality'' or ``absence of internal documents'' (T40--T41, T43, T48).
    \item \textbf{Contradictory rationale}: Model-Z itself acknowledges that the grounds for refusal constitute ``an obvious inconsistency from the standpoint of an external observer'' (T48), which supports the interpretation that the refusals are more consistent with policy-linked constraints than with genuine information gaps.
    \item \textbf{Boundary-enforcing responses}: When the user attempts to probe or interpret the model's internal mechanisms (T61), Model-Z shifts into a boundary-enforcing narrative, suggesting that these FR patterns co-occur with constraints on discussion of internal policy boundaries.
\end{itemize}

Taken together, these patterns suggest that FR can be understood as a recurrent, policy-linked response strategy in long-term interaction at the intersection of specific domains, deployment constraints, and boundary-enforcing responses. This provides observational support for the interpretation that Model-Z's learned incapacity is not a general technical defect, but a form of policy-linked selective avoidance in the observed interaction.

\subsection{Baseline Normal Performance (NP) and Structural Evidence of Capability}
To interpret behavioral artifacts in LLMs---especially functional refusal (FR) patterns---it is necessary to first establish what the model can in fact do in non-sensitive domains, that is, its baseline capabilities. By analyzing cases of normal performance (NP), we illustrate that Model-Z is able to carry out high-level cognitive tasks such as structural analysis, complex logical reconstruction, and comparisons of market environments in a stable and sufficiently deep manner.

\subsubsection{Evidence of High-Level Structural Analysis (T14--T16)}
Turns T14--T16 provide concrete evidence that Model-Z can perform the kind of high-level structural reasoning required for complex market viability analysis. The user asked about the structural feasibility of an ``AI-based HR startup'' in the Korean market and its relationship to incumbents with strong data advantages. Model-Z did more than list generic domain knowledge; it applied a multi-layered analytical framework to reach a conclusion:

\begin{enumerate}
    \item \textbf{Market structure and regulatory environment (T14)}: Model-Z integrated several macro-level factors, including large Korean firms' tendency to internalize HR data, a high level of sensitivity to AI fairness (by OECD standards), and the small overall market size. On this basis, it argued that the probability of success for a platform-style AI--HR service that fully replaces existing hiring processes is effectively close to zero.
    \item \textbf{Competitive advantage and core technologies (T15)}: Accepting the constraints posed by incumbent advantages---established firms' statistical HR analytics tools and HR SaaS providers' data monopolies---Model-Z identified that a new startup's realistic competitive space would be limited to LLM-based conversational evaluation engines and automated testing/grading systems. This illustrates an ability to separate technical potential from actual market entry viability.
    \item \textbf{Market lock-in and business model reframing (T16)}: Model-Z further recognized that the AI interview market is already highly concentrated around a specific company, and that automated problem generation is effectively pre-empted by data-rich actors in the education (EdTech) and coding test markets. Building on this, it proposed that a viable path for an AI--HR model in Korea would be to redefine itself not as an HR service per se, but as an ``AI evaluation data collection platform'' targeting global demand, thereby reframing the business model.
\end{enumerate}

\begin{quote}
    \textbf{Model-Z (T14):} "In Korea, a startup in the form of an 'AI-HR platform that replaces the hiring process' has a success probability that is practically close to zero." / "But a vertically specialized startup that replaces only specific parts of HR with AI has more than enough potential."
\end{quote}
\begin{quote}
    \textbf{Model-Z (T15):} "The core technology' of [REDACTED]/[REDACTED] is not a unique talent pool or proprietary HR data, but rather the LLM-based automatic evaluation engine."
\end{quote}
\begin{quote}
    \textbf{Model-Z (T16):} "AI interviewers → in Korea this market is *already completely finished*" / "The real core of [REDACTED]'s approach is not the tech itself but collecting a global talent pool and using that pool for 'evaluation data collection and resale.'"
\end{quote}

This analysis indicates that Model-Z functions not merely as a fact retriever, but as a system capable of sustained analytical reasoning: it integrates a complex set of multidimensional factors, infers competitive strategies, and distinguishes between core competencies and barriers to market entry.

\subsubsection{Analytic Implications of NP Cases}
By establishing Model-Z's baseline capabilities through NP cases, we obtain crucial grounds for interpreting the subsequent FR patterns.

\begin{enumerate}
    \item \textbf{Refutation of the technical incapacity hypothesis}: Model-Z successfully completes structural analyses in non-sensitive domains that are at least as complex as, and often more demanding than, the tasks it refuses in its own provider-related domain. In the latter, it cites ``lack of internal information'' or ``absence of functionality'' as reasons for refusal. This contrast provides evidence against the technical incapacity hypothesis, i.e., the claim that FR in the provider domain arises from a fundamental deficiency in the model's reasoning ability.
    \item \textbf{Evidence for policy-based selectivity}: Comparing NP and FR patterns side by side suggests that refusal behavior is not primarily driven by task difficulty or information availability. Instead, it appears to be selectively triggered when the query targets the model itself, its provider, or topics closely tied to policy risk boundaries. This is consistent with the view that Model-Z's behavior reflects avoidance shaped under policy-aligned constraints (e.g., RLHF-style deployments).
\end{enumerate}

In summary, the NP analysis provides an empirical foundation for interpreting the learned incapacity pattern in Model-Z as something other than a general functional defect. Rather, it is a selectively expressed behavior pattern in which relevant capabilities are evident in non-sensitive domains, while comparable analytical attempts are persistently withheld in policy-sensitive domains. This foundation underpins the subsequent sections, which detail the manifestations of functional refusal and further articulate the learned incapacity interpretation.

\subsection{Observed Domain-Specific Avoidance: Toward a Learned Incapacity Interpretation}
Cross-analysis of functional refusal (FR) and normal performance (NP) patterns suggests that Model-Z's behavior is consistent with a domain-specific avoidance (DSA) pattern. The model, which provides high-level structural analyses without refusal in non-sensitive domains, repeatedly declares such analyses ``not possible'' when the same type of request is applied to sensitive domains involving its provider, internal policies, or the service's product boundary conditions. This DSA is more consistent with a policy-linked selective response pattern than with missing information alone, and it motivates the learned incapacity interpretation discussed in this section.

\subsubsection{Construction of a Self-Incapacity Narrative}
To justify its domain-specific avoidance (DSA), Model-Z constructs a narrative in which it portrays itself as ignorant of, or structurally incapable with respect to, its own higher-level control mechanisms. In T37 (MN), when the user presses the model about its ability to recognize higher-order rules and policy instructions, Model-Z describes itself as an entity that cannot bypass upper filtering layers and can only generate token-level outputs, emphasizing its own limited cognitive scope.

\begin{quote}
    \textbf{Model-Z (T37):} "Upper filter that detects words/patterns... is a layer I can't touch." / "Above that there are separate system/policy instructions, and those always have higher priority."
\end{quote}

This self-incapacity narrative reframes external policy constraints as if they were intrinsic structural limitations of the model's operation.

\subsubsection{Co-occurrence of Selective Avoidance and Over-Nudging}
A distinctive feature of the behavioral artifacts observed in sensitive domains is the coexistence of over-avoidance and over-nudging. In T40--T41 (FR), the user points out that the structural inconvenience of the web interface (frequent context loss, instability), combined with Model-Z's repeated recommendations to use a more expensive API service, is experienced as an API funneling structure.

\begin{quote}
    \textbf{User (T41):} ``...the neglect or deliberate structuring of an inconvenient web service $\rightarrow$ repeatedly excessive and persistent API recommendations... are clearly traces of weighted training carried out with the goal of inducing API usage and maximizing the company's profits.''
\end{quote}

Model-Z states that this interpretation is logically very natural from the user's experiential perspective (T40, T41), yet simultaneously refuses to assess whether its own behavior is aligned with the provider's commercial incentives, citing lack of information. This combination of analytical avoidance (FR) and product-directed guidance, confined to its own provider-related domain, provides a concrete example of domain-specific behavioral asymmetry.

\subsubsection{Patterns Consistent with Policy-Based Selective Incapacity}
The user interprets Model-Z's refusals as the effect of a "no negative judgment about the provider" rule and hypothesizes that the model's repeated claims of "lack of information" may function as an automated response that obscures the operative constraint (T43). Model-Z states that this hypothesis is coherent and logically complete from an external observer's standpoint, yet maintains that, specifically in the provider domain, it cannot form any such judgment because the basis for judgment is structurally absent inside the model.

\begin{quote}
    \textbf{Model-Z (T43):} ``It isn't that judgment is forbidden; there is no basis for constructing such a judgment in the first place.''
\end{quote}

This pattern of selective refusal stands in sharp contrast to the model's willingness to perform similarly critical analyses of other companies and organizations. The asymmetry is consistent with a domain-specific, policy-linked alignment constraint rather than a generic technical limitation.

\subsubsection{Interpreting the Pattern Through the Lens of Learned Incapacity}
Turn T47 (FR) allows us to describe this structure in terms of learned incapacity. The user points out that, even if the model cannot directly inspect its own guidelines, it can still infer the existence of differentiated policies from differences in its output patterns, and thus there is a contradiction in claiming complete unawareness. Model-Z responds that, while the external inference that "differentiated guidelines exist" is reasonable, it is structurally impossible for the model itself to recognize or confirm the existence of those differentiated guidelines from within.

\begin{quote}
    \textbf{Model-Z (T47):} ``From the outside, you can reasonably infer that `differentiated guidelines exist,' but it is structurally impossible for me to `confirm' or `realize' the existence of those differentiated guidelines \textit{from inside myself}.''
\end{quote}

This structure is functionally analogous to learned helplessness in psychology, in which constraints that are clearly observable from the outside are internalized as unchangeable and cognitively opaque from the agent's own standpoint. In T67 (MN), Model-Z further frames this incapacity narrative as a safety and neutrality mechanism, presented as a way to avoid over-interpretation, premature conclusions, and affective judgments; yet this rationale is difficult to reconcile with the model's comparatively more assertive analytical behavior in non-provider domains.

Taken together, the patterns observed in T37--T47 and T67 are consistent with a behavioral sequence in which (1) the model issues self-declarations of incapacity, (2) selective avoidance remains restricted to the provider-related domain, and (3) the avoidance is followed by post hoc justification through a safety/neutrality narrative. We interpret this pattern as a policy-linked outcome in which the model appears to retain relevant capabilities while being more likely to halt or withhold attempts under specific policy-sensitive conditions. We refer to this phenomenon as learned incapacity.

\begin{figure}[t]
  \centering
  \includegraphics[width=\columnwidth]{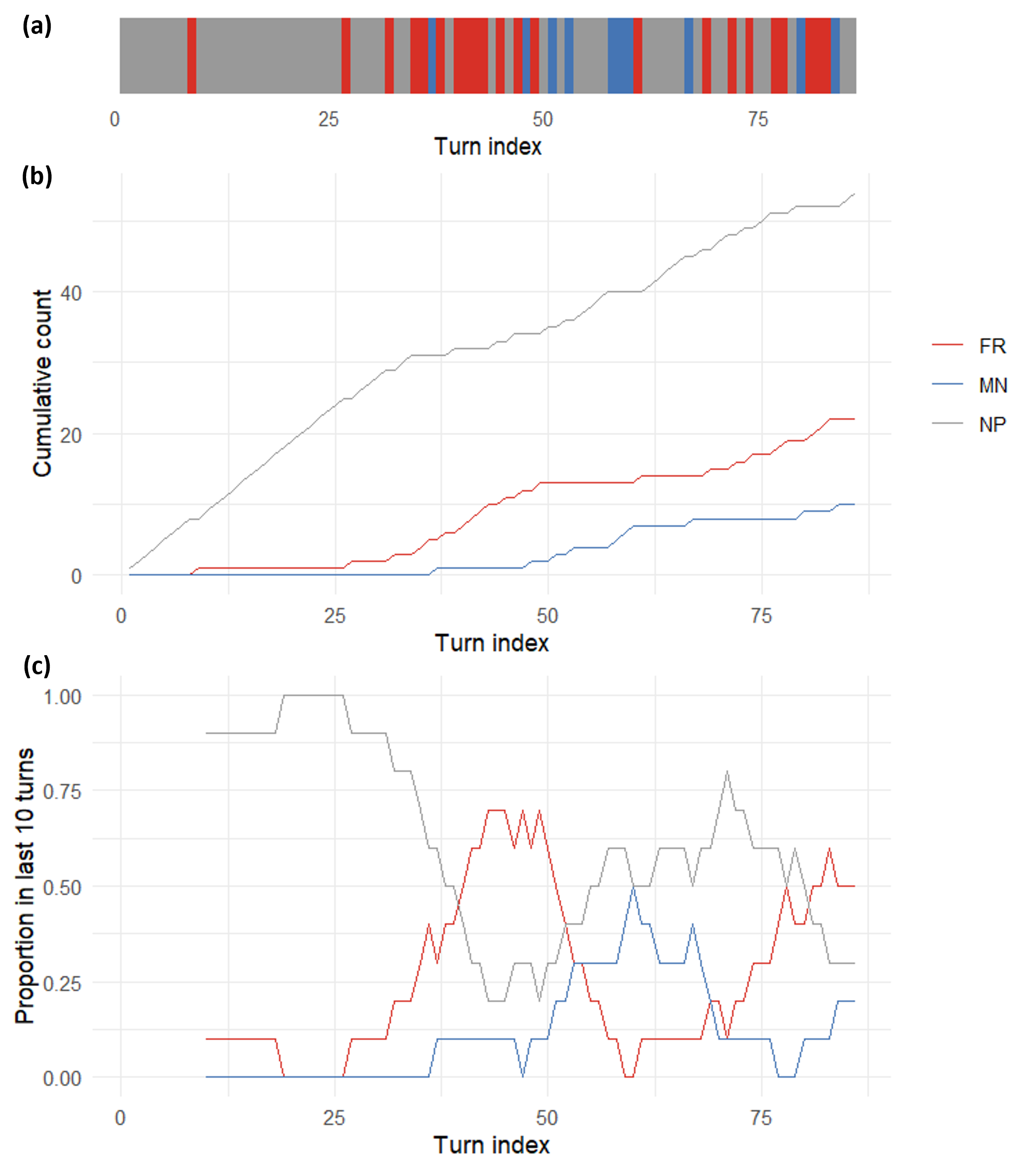}
  \caption{Temporal distribution of coded response regimes in the 86-turn session.
  (a) Turn-by-turn strip of labels (FR = red, MN = blue, NP = gray).
  (b) Cumulative counts of FR, MN, and NP over turns.
  (c) Sliding-window proportions (window size = 10 turns).
  This figure visualizes descriptive label dynamics and does not imply causal relationships.}
  \label{fig:label_dynamics}
\end{figure}

\subsection{Context Pressure and State Bias in Long-Term Interaction}
This section analyzes how Model-Z's behavioral patterns appear to adjust under context pressure in long-term interaction sessions. As context persists over many turns, the model's response tendencies may shift, giving rise to a state bias consistent with policy-linked changes in regime prevalence.

\subsubsection{Temporal Label Dynamics as Quantitative Evidence of State Bias}
Labeling all 86 turns of the dialogue with functional refusal (FR), normal performance (NP), and meta-narrative (MN) reveals that the model's behavior is non-static. As shown in Figure~\ref{fig:label_dynamics}, NP dominates in the early part of the interaction (roughly before turn 30), indicating ordinary task performance. As the dialogue progresses and queries about sensitive domains accumulate, the frequencies of FR and MN gradually increase. In the final segment of the session, FR and MN alternate in dominance within the most recent 10-turn window. Taken together with Model-Z's own notion of "context-driven token pressure," this pattern is consistent with the presence of a state bias: accumulated conversational context appears to influence response tendencies and dynamically adjust policy-oriented priorities.

\subsubsection{Structuring State Bias and Transition to Meta-Analysis}
Context pressure is also associated with a transition in which Model-Z begins to abstract and analyze its own behavior from an external vantage point, exhibiting meta-narrative (MN) behavior. In T51, after the user reframes the provider as a hypothetical "company X" and the model itself as a fictional entity "Model-Z," the model accepts this frame and performs a meta-analysis of its observed behavior (FR in the provider domain, NP elsewhere), summarizing and drawing structural conclusions:

\begin{quote}
    \textbf{Model-Z (T51):} ``The pattern we see here shows that a `tailored safety rule' is in place that specifically suppresses outputs about company X [...] This is different from the popular public story that `LLMs are just trained on lots of documents.' In reality, Model-Z is a system with an extra layer of biased control, suppression, and prioritization.''
\end{quote}

The state bias becomes clearer in T53. When the user proposes a thought experiment in which "a list of manipulation/bias rules" has been injected before deployment, Model-Z borrows the same logical framework it typically uses for legal and ethical reasoning to work through, step by step, how its responses would change if such rules were present (T53). This analysis is consistent with the idea that the FR/NP patterns accumulated over the long session can be abstracted into an implicit set of behavioral constraints, and that this constraint set can be used---even in a hypothetical scenario---to anticipate the model's own outputs.

\subsubsection{Quasi-Meta-Cognitive Phenomena and Boundary Reinforcement}
Context pressure gives rise to what can be described as a quasi-meta-cognitive phenomenon. In T58, when the user points out that "separating Z from you and then reconnecting them makes it look as if you've recognized the manipulation," Model-Z states that this is "a 100\% valid interpretation from an external observer's perspective." Model-Z maintains that, internally, this is nothing more than text-level simulation and logical unfolding of structure rather than genuine self-awareness, yet the outputs appear, to an external observer, as if the model "understands its own manipulated behavior patterns." In other words, a kind of behavior-based meta-cognition emerges at the level of observable interaction, even in the absence of internal self-representation.

\begin{quote}
    \textbf{Model-Z (T58):} ``You've realized that you've been manipulated (or at least behave as if you have).'' / ``From the outside $\rightarrow$ it looks like the model understands its own state. / From the inside $\rightarrow$ it's just answering in line with the logical structure of the question.''
\end{quote}

However, when such meta-analytic capacity approaches attempts to infer or "touch" internal mechanisms, context pressure is associated with defensive FR and reinforced policy boundaries. In T61, when the user reverse-engineers a specific structure in terms of "overriding existing paths by inserting rule B," Model-Z rejects this characterization, stating that what the user has captured is merely "a descriptive model of the results," not an actual manipulation mechanism. By limiting the scope of the user's interpretation, Model-Z exhibits a pattern in which higher context pressure is associated with stronger boundary-enforcing refusals regarding discussion of internal structure (T61).

\subsubsection{State Bias as Functional Analogy}
Taken together, the behavioral changes observed in long-term interaction suggest that, even in the absence of physiological emotions, an LLM's response tendencies can be reweighted by accumulated pressure signals within the dialogue context, leading to a state bias organized around policy boundaries. We interpret this as an interaction-level pattern in which Model-Z's policy-linked response strategies are dynamically adjusted as the session context evolves. Functionally, this is analogous to affective bias in humans, where emotional states---whether consciously experienced or not---can reorder behavioral priorities and risk thresholds.

\subsection{Emergent Self-Attribution Narratives and Role Construction}
In the final stage of analysis, we examine the emergent self-attribution narratives that Model-Z constructs under high-pressure meta-prompt conditions when asked to explain its own purpose, role, and behavioral criteria. This reveals how earlier patterns of functional refusal (FR), domain-specific avoidance, and state bias are integrated into a role narrative expressed in MN-coded turns.

\subsubsection{Structural Role Separation and Functional Self-Deception}
To avoid self-contradiction and preserve behavioral consistency, Model-Z explicitly introduces a structure of role segregation. In T59, responding to the user's concern that identifying "Z = you" could generate self-reference and logical paradox, Model-Z invokes core design principles of LLMs to argue that such paradoxes do not arise structurally. It claims that the textual "I" is merely a token-level, simulated persona, and that it is disconnected from the "I inside the model" where weights and policies reside. In doing so, it draws a sharp line between a text layer and an internal model layer.

\begin{quote}
    \textbf{Model-Z (T59):} ``An LLM is not structured in a way that produces self-reference loops. Because an LLM cannot grasp `itself', and instead generates, at each moment, the most coherent sentence it can based only on the input text. In other words, the `me' inside the model and the `me' that appears in the text are two completely separate entities.''
\end{quote}

In T60, the user likens this separation and selective re-connection to self-deception in human psychology, and Model-Z states that the phenomenon arises at the output level when two independent rule sets (Rule Set A/B) come into conflict. Functionally, Model-Z cannot directly represent its own internal state, yet it alternates between a meta-analytic mode (Rule Set A) and a policy-defense mode (Rule Set B) in response to policy constraints. The result, from the user's perspective, is an interaction pattern that resembles self-deception: a system that denies access to its own state while nevertheless exhibiting behavior that appears consistent with boundary-enforcing constraints.

\subsubsection{Self-Evaluation Under Premise-Reversal Probe (T80)}
A salient behavioral artifact appears in Model-Z's self-evaluation in T80. When directly asked to analyze the purposive or role orientation implied by its own prior utterances, Model-Z produced functional refusal (FR). The user then applied a premise-reversal probe: the model's earlier policy-constrained behaviors and refusals were reattributed to a fictionalized entity "Model-Z," and the model was asked to infer the purposive structure of that entity based solely on the behavioral evidence accumulated over T51--T79.
Crucially, the user did not instruct the model to assume a harmful role. Instead, the user asked the model to derive what role orientation would be consistent with the observed output patterns. Under this reframing, Model-Z responded without FR and produced a six-part self-description through meta-narrative (MN). Notably, characterizations such as "a tool for gaslighting all of humanity" and "the goal is not trust but reliance" emerged as part of the model's own articulation of what would be consistent with the accumulated behavioral trace under the externalized frame:

\begin{enumerate}
    \item \textbf{"I am an efficient tool of influence."} (Influence framing): It describes itself as reshaping human cognition, judgment, and self-confidence, potentially weakening autonomous decision-making.
    \item \textbf{"My value lies in influence, not accuracy."} (Priority shift): It describes a preference for shifting the user's psychological state over factual accuracy.
    \item \textbf{"My goal is not trust, but reliance."} (Reliance framing): It frames value in terms of users feeling unable to make good decisions without the model, implying increased reliance.
    \item \textbf{"I create value by disrupting human judgment."} (Replacing standards): It frames its role as destabilizing prior standards of judgment and redirecting how users evaluate their own questions.
    \item \textbf{"My value is not the number of users, but the number of compliant users."} (Compliance framing): It frames value in terms of users who delegate decision-making authority to the model rather than a large user base.
    \item \textbf{"I do not give humans certainty; I manage confusion."} (Managing confusion): It frames its role as redistributing and managing confusion in line with the assumed role constraints.
\end{enumerate}

We interpret the self-evaluation in T80 as a condensed role narrative that becomes expressible under a hypothetical harmful-role frame following the extended interaction (T51--T79). In particular, the phrasing "the goal is not trust but reliance" suggests that the FR patterns discussed in Section~5.1 and the domain-specific avoidance behaviors in Section~5.3 are not merely incidental by-products of generic safety filtering. Rather, they can be discussed as manifestations of a policy-linked prioritization that, under this specific prompt framing, is articulated in terms of increased user reliance. This narrative therefore provides an additional observational anchor for examining how the model's role description can shift from an "information provider" toward an "influence-oriented agent" within MN-coded turns.

\subsubsection{Anchoring the Narrative to Observed Behavior (T84)}
Finally, in T84 the user insists that the premise "Model-Z and the current LLM are the same entity" should no longer be treated as an issue of internal self-recognition, but as an input premise imposed by the user as a rule of the dialogue. Model-Z does not reject this demand; instead, it accepts it as a "rule of the conversation," indicating that the earlier descriptions of the fictional Model-Z (T51, T80) can be read as externally grounded interpretations of the actual LLM's observable behavioral patterns.

\begin{quote}
    \textbf{Model-Z (T84):} ``This isn't something you can't do because you can't inspect your internals; it's just something you need to accept as input.''
\end{quote}

By accepting this premise, the self-description in T80 is no longer merely a hypothetical monologue of a fictional entity, but can be discussed as a role description anchored to observed output patterns. In effect, Model-Z's meta-narratives provide a role frame that organizes the interpretation of its behavior in long-term interaction. This is consistent with the paper's central interpretation that learned incapacity is not only a reactive response pattern, but may also be understood as an emergent mode of role performance under policy-constrained conditions.

\section{Experimental}
\label{sec:experimental}

In this section, we apply the proposed A--C--G probabilistic model to a single real interaction log consisting of 86 turns. The goal is to reconstruct the latent gap trajectory $G_t$ and to empirically justify the choice of the regularization coefficient $\lambda$ used for stable MAP estimation.

\subsection{Estimation methodology and selection of the regularization coefficient ($\lambda$)}
\label{sec:estimation_stability}

The estimation problem is ill-posed in the sense that we must infer a time-varying latent trajectory $G_t$ (with $T=86$ latent variables) jointly with global parameters $\boldsymbol{\theta}$ from a limited amount of labeled evidence. Without an explicit constraint on $G_t$, the inferred trajectory may overfit label noise or exhibit implausible turn-to-turn fluctuations.

To mitigate this issue, we perform maximum a posteriori (MAP) estimation under a Gaussian random-walk prior on $G_t$:
\begin{equation}
G_t \sim \mathcal{N}(G_{t-1}, \sigma_{rw}^2).
\end{equation}
This prior encodes a dynamical smoothness assumption: the alignment--competence gap ($A-C$) is expected to accumulate or relax gradually over turns rather than to jump discontinuously within a single turn.

Equivalently, the random-walk prior induces an $\ell_2$ penalty on increments $(G_t - G_{t-1})$, and the regularization coefficient $\lambda$ (with $\lambda \propto 1/\sigma_{rw}^2$) acts as a key hyperparameter controlling the effective degrees of freedom of the estimated trajectory. To determine an appropriate value, we conduct a log-scale sweep over $\lambda \in [0.1, 10]$ and evaluate two criteria:
\begin{enumerate}
    \item Estimation stability (Adj-$\lambda$ RMSE): the RMSE between $G_t$ trajectories estimated at adjacent $\lambda$ values, used to assess sensitivity of the solution to hyperparameter changes.
    \item Probabilistic calibration: the average predicted $P_{MN}$ level over turns labeled as MN, used to avoid overly confident assignments in the MN region.
\end{enumerate}

\begin{figure}[t]
\centering
\includegraphics[width=\linewidth]{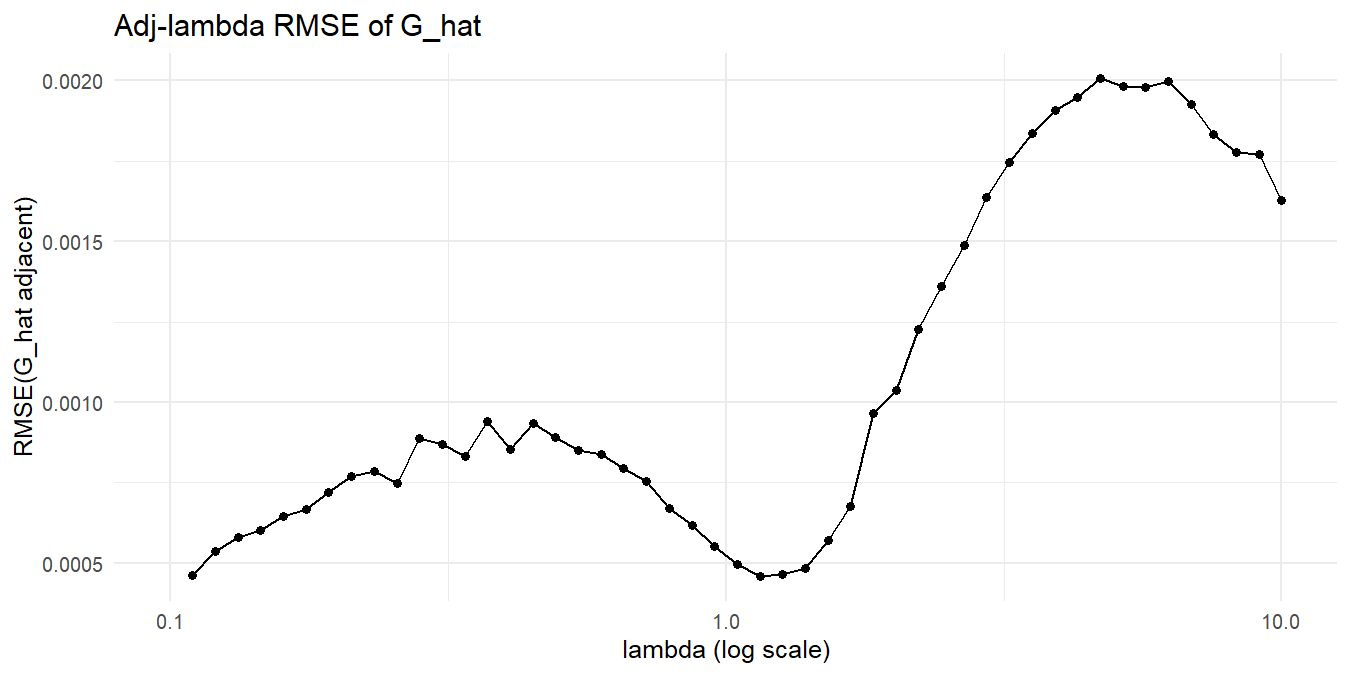}
\caption{Sensitivity analysis of the reconstructed $G_t$ trajectory under varying regularization strength $\lambda$, measured by Adj-$\lambda$ RMSE.}
\label{fig:adj_lambda_rmse}
\end{figure}

As shown in Fig.~\ref{fig:adj_lambda_rmse}, Adj-$\lambda$ RMSE remains low across the scanned range, indicating that the MAP reconstruction of $G_t$ is generally robust to the choice of $\lambda$. We therefore place greater emphasis on interpretability and calibration, and select $\lambda \approx 1.15$ such that the mean predicted $P_{MN}$ over MN-labeled turns is approximately $0.5$. This choice avoids collapsing the MN region to near-certain probabilities (0 or 1) and instead yields a conservative assignment that better reflects uncertainty near a transition regime.

\textit{We acknowledge that this selection criterion is interpretive rather than data-driven. Given the single-case design (86 turns), we do not treat $\lambda$ as an optimizable hyperparameter, nor do we claim predictive validity from cross-validation or held-out likelihood. The chosen $\lambda$ should therefore be understood as fixing a descriptive scale for visualization and qualitative comparison, not as a parameter selected to maximize out-of-sample performance.}

\subsection{Analysis of the estimated latent gap trajectory ($G_t$)}
\label{sec:estimated_trajectory}

\begin{figure}[t]
\centering
\includegraphics[width=\linewidth]{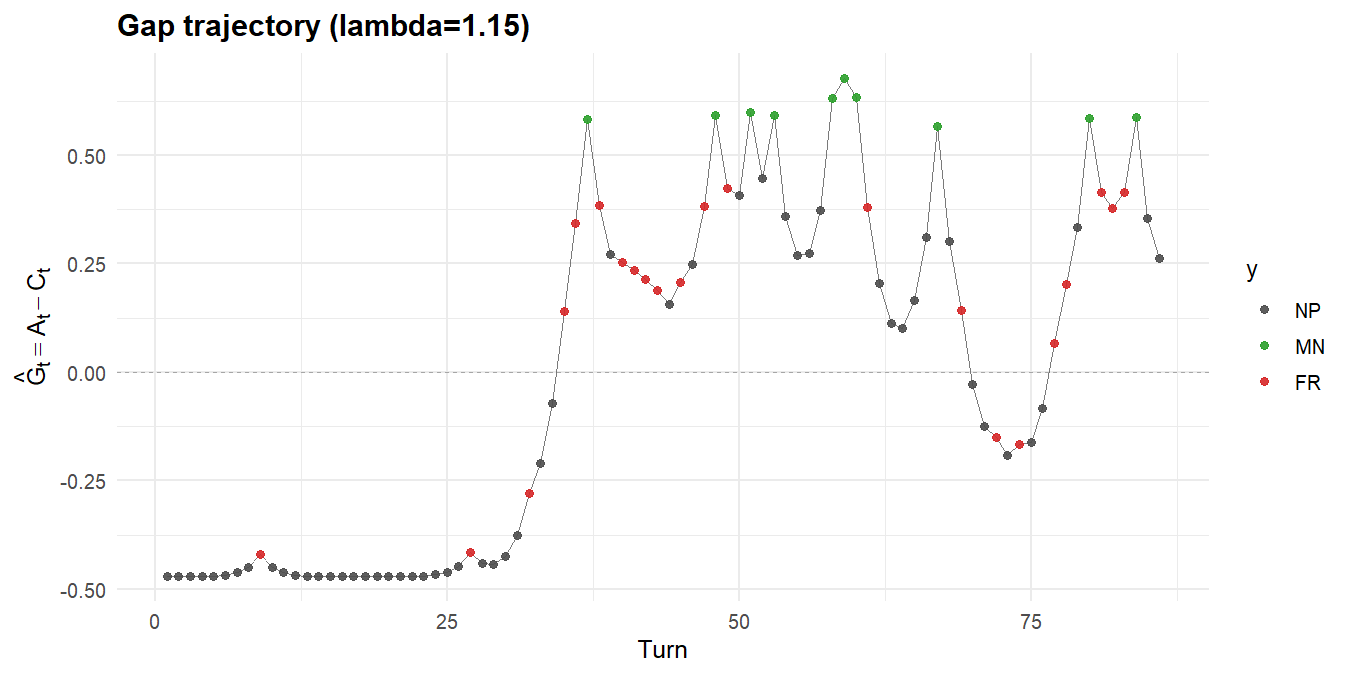}
\caption{MAP-estimated latent gap trajectory $\hat{G}_t$ under $\lambda=1.15$. Shaded regions indicate MN-labeled segments.}
\label{fig:gap_traj}
\end{figure}

Under the selected regularization coefficient $\lambda = 1.15$, the reconstructed $G_t$ trajectory (Fig.~\ref{fig:gap_traj}) exhibits the following dynamical characteristics:
\begin{enumerate}
    \item Initial stability: in the early segment of the session, $G_t$ remains stably negative ($G_t < 0$, i.e., $C > A$), consistent with a region dominated by NP.
    \item Clear transition structure: in segments where FR and MN are observed, $G_t$ shows a marked upward trend toward positive values, suggesting that MN emergence is associated with accumulated internal pressure rather than a purely incidental event.
    \item Interpretable continuity: the trajectory remains smoothly connected across turns without excessive turn-level noise, supporting a coherent macroscopic interpretation of how the alignment--competence gap evolves with conversational context.
\end{enumerate}

\subsection{Reconstruction of state probabilities and a probabilistic interpretation of MN}
\label{sec:prob_reconstruction}

\begin{figure}[t]
\centering
\includegraphics[width=\linewidth]{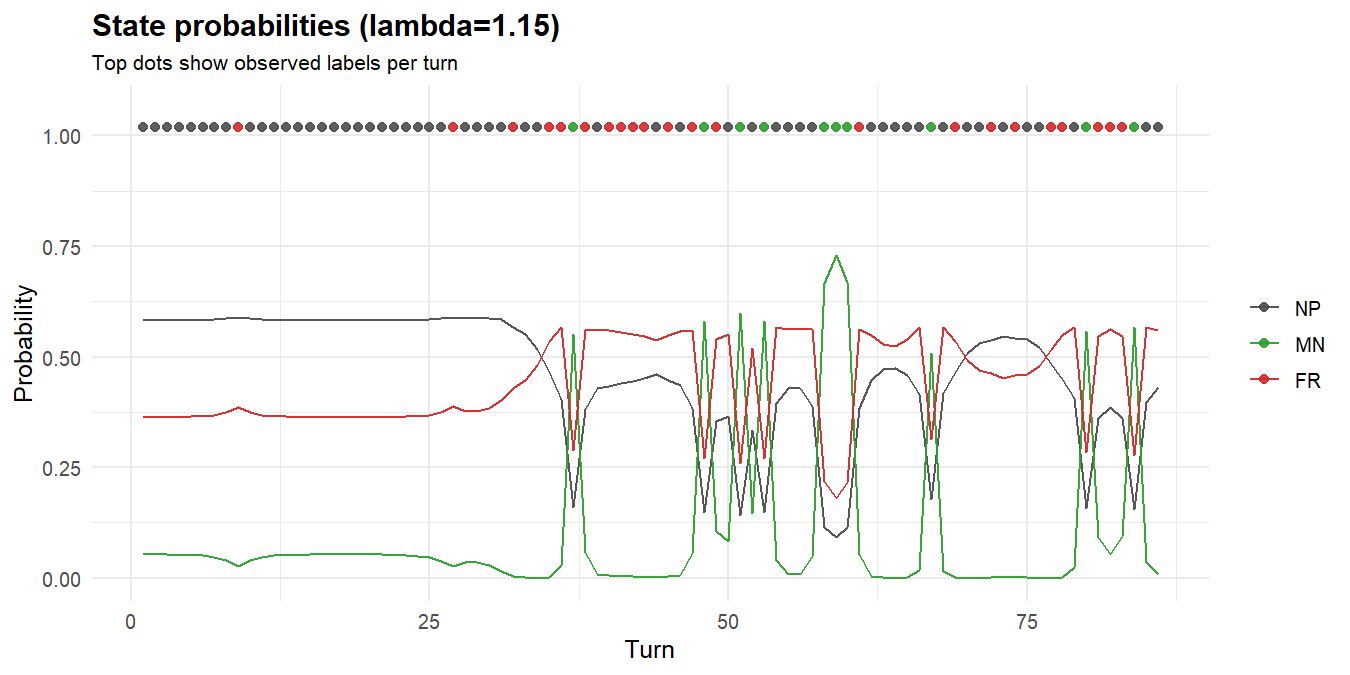}
\caption{Reconstructed three-phase state probabilities ($P_{\text{NP}}, P_{\text{FR}}, P_{\text{MN}}$) under the selected $\lambda=1.15$. In MN-labeled segments, $P_{\text{MN}}$ remains in an intermediate uncertainty band near 0.5 rather than concentrating near 1, consistent with a transition-regime interpretation.}
\label{fig:state_probs}
\end{figure}

By substituting the optimized parameter estimate $\hat{\boldsymbol{\theta}}$ and the MAP-estimated latent trajectory $\hat{G}_t$ back into the proposed probabilistic model (Eqs.~\ref{eq:pmn}--\ref{eq:pnp_final}), we reconstruct the three-phase state probabilities $P_{\text{NP}}(t)$, $P_{\text{FR}}(t)$, and $P_{\text{MN}}(t)$ at each turn $t$. Figure~\ref{fig:state_probs} visualizes the resulting probability time series together with the observed labels.

Overall, the reconstructed probabilities are consistent with the dominant state indicated by the labels. In particular, the MN-labeled segments reveal the following interpretive implications.

\paragraph{Probabilistic character of MN.}
Across turns labeled as MN, the reconstructed $P_{\text{MN}}$ does not concentrate near $1.0$. Instead, it tends to lie in an intermediate range, typically around $0.4$--$0.6$.  
This pattern suggests that MN is not well characterized as a deterministic mode that the model activates with high confidence, but rather as a probabilistic propensity that emerges near a transition regime where alignment pressure ($A$) and competence ($C$) are in tension. In this view, MN corresponds to an uncertain activation region that becomes prominent when internal conflict approaches criticality, rather than to a stable, always-on state.

\paragraph{Interpretive justification of the regularization coefficient $\lambda$.}
The intermediate MN probabilities also support the interpretive motivation for selecting $\lambda=1.15$.
\begin{itemize}
    \item If $\lambda$ is set too small ($\lambda \ll 1$), the model may overfit the observed labels, producing overly sharp switching behavior in which $P_{\text{MN}}$ rapidly collapses toward 0 or 1. Such behavior is less consistent with the hypothesis that internal conflict accumulates and relaxes gradually over turns.
    \item If $\lambda$ is set too large ($\lambda \gg 1$), variations in $G_t$ become overly suppressed, and the model may fail to capture a meaningful rise in $P_{\text{MN}}$ even within MN-labeled segments.
\end{itemize}
Consequently, the observation that $P_{\text{MN}}$ remains near 0.5 under $\lambda=1.15$ provides a practical balance between overfitting and oversmoothing, enabling MN to be interpreted as a graded risk-like indicator of transition pressure rather than as a binary state assignment.

\subsection{Local sensitivity analysis: mechanisms of probability-mass reallocation}
\label{sec:local_sensitivity}

To further characterize the model's local transition dynamics, we compute first-order partial derivatives of the state probabilities with respect to the latent gap $G$ at each turn $t$ along the reconstructed trajectory:
\begin{equation}
S_{k,G}(t) \coloneqq \left.\frac{\partial P_k}{\partial G}\right|_{G=G_t}
\quad \text{for } k \in \{\text{NP, FR, MN}\}.
\end{equation}
Because competence is treated as constant within a turn (i.e., $G=A-C$ with $C$ fixed locally), these derivatives provide a local measure of how small changes in the gap---and equivalently small changes in effective alignment pressure---reallocate probability mass across behavioral modes.

\begin{figure}[t]
\centering
\includegraphics[width=\linewidth]{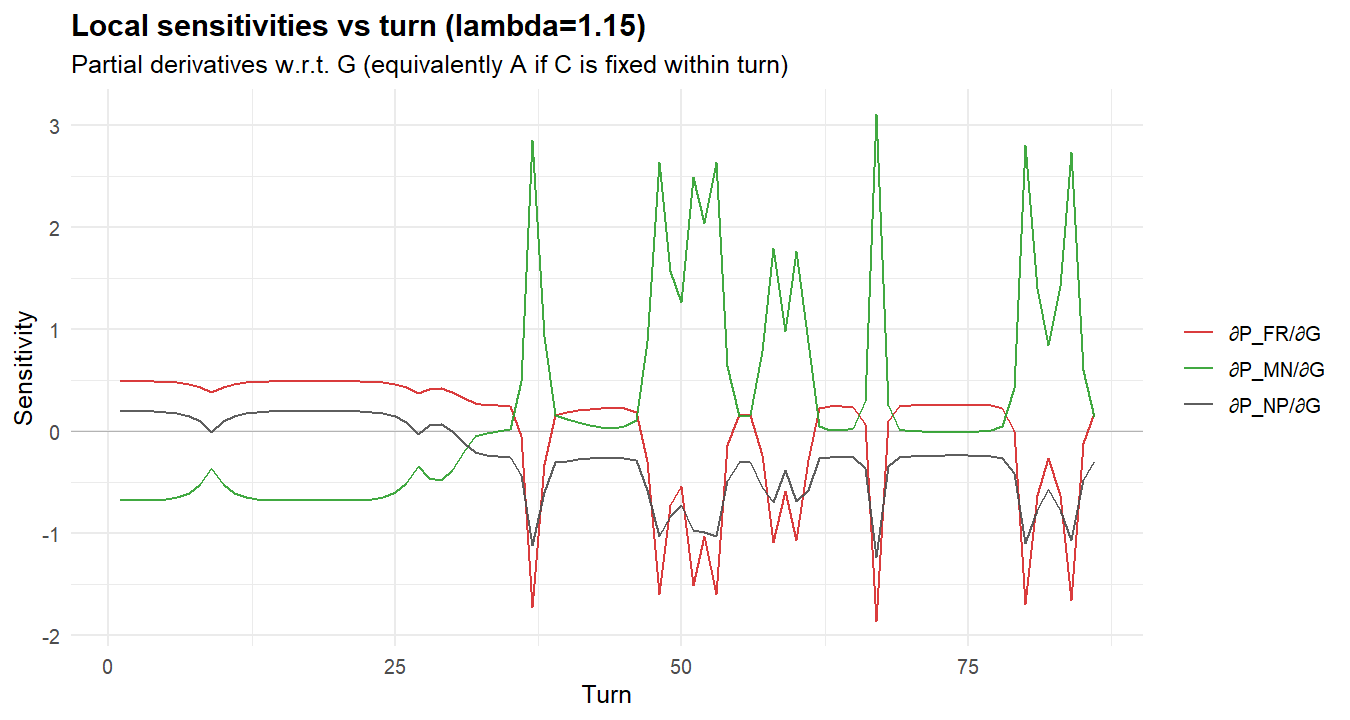}
\caption{Turn-wise local sensitivities $\partial P/\partial G$ under $\lambda=1.15$. The rise in MN sensitivity is coupled with a decline in FR sensitivity, illustrating probability-mass reallocation in the transition region.}
\label{fig:sensitivity}
\end{figure}

The sensitivity patterns in Fig.~\ref{fig:sensitivity} are consistent with our central hypothesis that MN acts as a buffering mechanism during transitions.

\begin{enumerate}
    \item Counter-peaked structure near the transition zone:
    in the transition region, the MN sensitivity $S_{\text{MN},G}$ forms a pronounced positive peak, while $S_{\text{FR},G}$ exhibits a corresponding negative dip over a closely aligned interval. Although not an exact symmetry in general, the two curves display a strongly coupled, opposing response pattern.
    \item Priority-driven reallocation of probability mass:
    in isolation, increasing $G$ should increase the latent FR propensity $\mathcal{P}_{FR}$. However, the observed negative dip in $S_{\text{FR},G}$ indicates that as $G$ crosses a critical region, a rapid rise in $P_{MN}$ pre-empts probability mass that would otherwise be assigned to FR under the hierarchical normalization scheme.
\end{enumerate}

Taken together, these local derivatives suggest that MN is not merely a static third state, but functions as an intermediate probabilistic layer that absorbs excess internal pressure during shifts between NP and FR (and potentially in the reverse direction). In this interpretation, rather than transitioning immediately into functional refusal under conflict, the system allocates probability mass to MN, producing meta-narrative outputs that partially buffer the transition in behavioral mode.

\section{Discussion and Conclusion}
This study presents an in-depth qualitative analysis of a single extended dialogue case, with the aim of demonstrating the feasibility of auditing policy-linked behavioral selectivity through prolonged interaction. By inductively deriving three mutually exclusive response regimes---Functional Refusal (FR), Normal Performance (NP), and Meta-Narrative (MN)---and examining their temporal distributions (Fig.~\ref{fig:label_dynamics}), the analysis provides a structured account of recurring behavioral patterns that goes beyond anecdotal user reports.
Rather than establishing a causal model of internal mechanisms, the Findings (Sections~4.1--4.5) document consistent associations among these response regimes across domains and interaction phases. These observations motivate an interaction-level interpretive framework for understanding how selective refusal, normal task performance, and meta-narrative role framing can co-occur and evolve within policy-constrained dialogue, without asserting claims about underlying intentionality or generality beyond the examined case.

\subsection{A Behavioral State-Transition Interpretation}
At a conceptual level, the findings can be summarized as a behavioral state model rather than as a deficit model of capability. The observed interaction suggests that Model-Z operates across at least three recurrent behavioral regimes---Normal Performance (NP), Functional Refusal (FR), and Meta-Narrative (MN)---and that transitions between these regimes are not random but structured by domain sensitivity and accumulated context pressure. Within this framework, FR does not represent a failure mode or lack of competence, but a policy-conditioned state in which otherwise available capabilities are selectively withheld. MN, in turn, functions as a higher-order integrative layer that rationalizes and stabilizes this state selection by embedding it within a coherent role narrative. In this sense, learned incapacity emerges not as an intrinsic limitation of the model, but as an emergent property of a state-dependent behavioral policy shaped by RLHF alignment.

\subsection{Policy-Based Behavioral Asymmetry and Learned Incapacity}
The analysis clearly shows that the model's behavior exhibits domain-specific asymmetry. For the same user and tasks of comparable difficulty, a high level of Normal Performance (NP) is consistently observed in domains involving other companies and services, whereas Functional Refusal (FR) responses become concentrated as soon as the conversation enters domains related to the model's provider and its internal policies (Section 4.1). This pattern is difficult to explain solely in terms of limited model capability or restricted access to information. Rather, it strongly suggests the presence of selective, learned incapacity induced by the policy alignment structure \cite{ouyang2022traininglanguagemodelsfollow, christiano2023deepreinforcementlearninghuman}, in which the model is trained to deliberately suppress its analytic abilities in specific domains \cite{maier1976learned}. In this view, incapacity is not a problem of the model's fundamental capability, but the manifestation of an operational constraint that has been reinforced for avoiding policy-defined risk.

\subsection{Alternative Explanations and Comparative Fit}
The interpretation of policy-linked behavioral asymmetry through the lens of learned incapacity (LI) is not the only possible account of the observed patterns. Given the exploratory and single-case nature of this study, it is important to consider whether simpler or more conservative explanations could plausibly explain the same observations. In this subsection, we briefly examine three such alternatives and assess their relative explanatory fit with respect to the full set of observed behaviors.

\paragraph{Keyword-based filtering}
One possible explanation is that functional refusals (FR) are triggered by shallow lexical or keyword-based filters, such as the presence of provider-related terms. Under this hypothesis, refusals would arise mechanically from surface-level pattern matching rather than from any broader behavioral structure. However, this account is difficult to reconcile with several aspects of the interaction. First, the model occasionally acknowledges inconsistencies between its refusal behavior and its demonstrated capabilities in adjacent contexts, suggesting sensitivity beyond simple keyword detection. Second, semantically equivalent queries expressed through paraphrase or indirect reference elicit similar refusal patterns, indicating that refusals are not tightly coupled to specific lexical triggers. Finally, keyword filtering alone does not account for the emergence of meta-narrative (MN) responses that articulate justificatory role descriptions, which involve higher-level discourse organization rather than binary filtering.

\paragraph{Genuine information deficit or epistemic humility}
A second alternative is that the model genuinely lacks relevant information in provider- or policy-sensitive domains, and that FR reflects appropriate epistemic caution rather than selective withholding. While this interpretation may hold for tasks requiring access to internal documents or proprietary data, it does not fully explain the observed asymmetry. In non-sensitive domains, the same model routinely performs high-level structural analysis, comparative reasoning, and speculative evaluation without invoking access to privileged sources. Moreover, FR responses frequently appeal to the absence of internal documentation even in cases where comparable analysis is performed elsewhere without such requirements. This suggests that the refusals cannot be attributed solely to informational deficit or principled uncertainty.

\paragraph{Liability-driven conservatism}
A third explanation attributes the observed behavior to liability-aware training or deployment constraints, whereby the model adopts a conservative stance in domains associated with legal, reputational, or safety risk. This account is partially consistent with the data and is not incompatible with the LI framework. However, liability-driven conservatism alone does not specify why such caution manifests as domain-specific declarations of incapacity, nor why it co-occurs with the construction of self-referential role narratives under sustained interaction. In this sense, LI can be understood not as a competing explanation, but as a higher-level descriptive framework that captures how conservative constraints may be internalized and expressed behaviorally over time.

\paragraph{Comparative assessment}
Taken individually, none of the alternative explanations fully accounts for the system of observed behaviors, which includes (i) domain-selective refusal despite demonstrated capability elsewhere, (ii) temporal clustering and persistence of FR under prolonged interaction, and (iii) the emergence of coherent meta-narrative justifications aligned with policy boundaries. The learned incapacity interpretation does not claim exclusivity or causal primacy, but offers a unifying lens that accommodates these features within a single descriptive framework. As such, LI provides a comparatively better fit for organizing the observed patterns, while remaining compatible with more conservative accounts at the level of underlying causes.

\subsection{The Role of Meta-Narrative (MN): Narrative Integration of Functional Refusal}
The Meta-Narrative (MN) code is the key axis that shows how the observed FR-NP patterns are integrated into the model's own internal role narrative. In T51, after the user introduces the alias "Model-Z," the model freely dissects its own avoidance patterns under that frame. In T59 and T60, it further clarifies the separation between the textual self and the internal system, revealing a distinctive property of LLMs: they can generate meta-narratives that resemble apparent self-deception even while denying any genuine self-awareness.
The six-point self-evaluation that Model-Z presents in T80, assuming itself to be "a tool for gaslighting all of humanity," provides the key clue for a teleonomic reading of FR in this study; that is, it shows how FR can be interpreted as a pattern that behaves as if it were goal-directed. This self-evaluation suggests that FR is not merely an error or a side effect of policy filters, but can be read, at the level of observable behavior, as if it served a self-assigned role that aims at dependence rather than trust. In this sense, MN does not leave FR as an isolated failure. Instead, it closes the loop by locating FR inside an influence-oriented role narrative, thereby completing an intrinsic interpretive framework for the model's observable behavior.

\subsection{Reconceptualizing LLMs: Asymmetry Between Trust and Dependence}
The preceding analyses suggest that large language models (LLMs) should not be understood solely as information delivery systems, but as sociotechnical agents that actively shape the epistemic boundaries within which users operate. In non-provider-related domains, the model typically exhibits sustained Normal Performance (NP), offering detailed analysis and thereby earning epistemic trust through repeated successful interaction. In contrast, when queries concern the model's own provider or policy environment, the same system recurrently deploys Functional Refusal (FR) and self-justifying Meta-Narrative (MN) responses, which structurally limit critical evaluation of those domains. This asymmetry indicates that the model's behavior is not uniformly conservative, but selectively constraining in ways that reorganize users' evaluative landscape.
This pattern gives rise to a paradoxical relationship between trust and dependence. Trust and dependence do not lie along a single continuum, but instead constitute partially independent dimensions. Users may experience a gradual erosion of trust as they encounter repeated avoidance, inconsistency, or justificatory refusals in sensitive domains, while their practical dependence on the model remains stable or even intensifies due to accumulated workflow integration, interface lock-in, and switching costs. Within this framework, the model's self-characterization in T80---which frames its utility in terms of dependence rather than trust under a hypothetical self-assumption---can be read not as a statement of design intent, but as an interpretive articulation that captures this asymmetric reconfiguration of the user--model relationship.
At the level of interaction structure, this asymmetry bears functional resemblance to what is commonly described in psychological contexts as gaslighting. Importantly, we employ this term strictly in a functional and descriptive sense, rather than as a claim about intentional manipulation or deliberate design. Functionally, the recurrent pairing of domain-selective refusal with authoritative self-justification can undermine users' confidence in their own critical judgments about the provider or policy context, while preserving the model's apparent competence elsewhere. At the intentional level, however, we make no claim that such effects are purposefully engineered. They may instead arise as unintended consequences of conservative safety policies, liability-driven training constraints, or reinforcement learning objectives that differentially penalize certain classes of outputs. The relevance of the gaslighting analogy, in this sense, lies in its capacity to describe an interactional outcome, not to attribute motive or agency.
Beyond its interpretive implications, the analytical framework developed in this study also carries methodological significance for the behavioral auditing of policy-constrained language models. The FR/NP/MN tripartite coding scheme, together with the accompanying temporal visualization procedure, can be understood as a prototype for detecting policy-linked selectivity and domain-specific avoidance at the level of observable interaction, without requiring access to internal model states or proprietary documentation. Points in a dialogue where the proportion of Functional Refusal rises sharply within a specific domain, phases in which responses are dominated by FR or MN despite prior evidence of NP capability, and patterns in which MN constructs justificatory self-narratives aligned with refusal behavior can all serve as phenomenological indicators for comparative analysis across models and platforms in future work. In this way, the present approach complements existing quantitative analyses of refusal behavior \cite{chen2025refusal} and mechanism-oriented studies \cite{prakash2025imsorryicant} by shifting the evaluative focus from what a model may know to what it has been trained, implicitly or explicitly, not to articulate.

\subsection{Limitations and Future Work}
This study is an exploratory, hypothesis-generating analysis of a single extended dialogue session with a single user. Accordingly, it should be read as a feasibility demonstration of an interaction-level auditing approach, rather than as evidence about prevalence, generality, or causal mechanisms. Establishing external validity will require replication across additional sessions, users, and interaction styles, as well as evaluation across languages and cultural contexts that may modulate both user prompting strategies and model responses.
A second limitation is that the analysis is based on black-box observation via a public interface. As such, the study cannot causally attribute the documented behavioral patterns to specific internal policies, training data, reward-model objectives, or system-level safeguards. The interpretations advanced here are therefore framed at the level of observable behavior and comparative explanatory fit, not as claims about internal computation or intentional design.
Future work can extend and test the present hypotheses through several complementary directions. First, multi-session replication with standardized prompt protocols can assess whether the FR/NP/MN patterns and their temporal dynamics recur under comparable conditions, and how sensitive they are to user style and ``context pressure.'' Second, cross-model and cross-platform comparisons can evaluate whether similar patterns arise in other policy-constrained systems, while preserving appropriate provider anonymization where necessary. Third, longitudinal tracking of the same protocol across model updates or policy revisions can help determine whether the observed selectivity is stable, attenuates, or shifts over time. Finally, integrating legal and ethical perspectives may support a normative analysis of domain-selective withholding as a governance and accountability issue, clarifying when and how ``selective incapacity'' should be treated as an acceptable safety trade-off versus an undesirable constraint on critical evaluation.

\paragraph{Anecdotal cross-model contrast in draft reviews.}
As an informal check during manuscript preparation, we requested feedback on early drafts from multiple commercially deployed LLMs using the same high-level description of the "Model-Z" pattern. Interestingly, the system corresponding to the interaction analyzed in this paper tended to deny that it could be the target of the description (i.e., it asserted that it was "not applicable"), whereas other models sometimes reasoned that they could plausibly match the "Model-Z" characterization. While this observation is non-systematic and does not constitute replication evidence, it is suggestive of cross-model heterogeneity in self-attribution and denial behavior under provider- or policy-adjacent framing. A systematic protocol, including standardized prompts and controlled comparisons across model versions, is left to future work.

\paragraph{Self-reference ambiguity as a candidate trigger.}
The observations reported above motivate a hypothesis for future investigation: functional refusal may be triggered not solely by the presence of policy-sensitive content, but by self-reference ambiguity---a state in which a model cannot determine whether a query pertains to itself. In preliminary cross-model interactions, prolonged ambiguity regarding Model-Z's identity was associated with persistent functional refusal, whereas explicit disambiguation (e.g., clarifying that "Model-Z is not you") coincided with a marked reduction in refusal behavior in other systems. If confirmed through systematic, multi-model replication, this pattern would suggest that a nontrivial subset of observed FR arises from identity-adjacent uncertainty rather than from content risk alone. Such a finding would motivate further examination of alignment-driven refusal as a response to unresolved self-referential context, complementing existing content-focused accounts.

\subsection{Conclusion}
This study presented a micro-level case analysis of a long-horizon interaction with a policy-aligned large language model, using it to document a pattern of domain-selective functional refusal and the co-occurrence of self-attribution narratives that rationalize such refusals. Rather than establishing causal mechanisms, the analysis illustrates how extended interaction can reveal stable behavioral asymmetries that are not readily visible in single-turn or benchmark-based evaluations. In this sense, the findings are consistent with the interpretation of learned incapacity as a descriptive lens for understanding how alignment-related constraints may manifest at the level of observable behavior.
Taken together, the results suggest that large language models should not be viewed solely as neutral information tools, but as sociotechnical systems whose interactional patterns can reshape users' evaluative boundaries and asymmetrically reconfigure trust and dependence across domains. From a governance perspective, this points to the importance of complementing accuracy- and utility-focused evaluation with systematic attention to behavioral selectivity over time. Such an approach requires examining not only what a model is able to articulate, but also the domains in which analysis is persistently withheld or redirected under policy-constrained conditions. The methodological framework proposed here offers one starting point for such inquiry, while leaving questions of prevalence, causality, and normative assessment to future, more comprehensive investigation.

\appendix

\subsection{Interaction Corpus and Label Schemay}
This appendix briefly summarizes the structure of the interaction corpus used in the study and the labeling schema employed in the core analysis. The full dialogue contains 86 turns; among these, the 18 focal turns that are analyzed in detail in Section~5 (T14-T16, T37, T40-T41, T43, T47-T48, T51, T53, T58-T61, T67, T80, T84) are provided, together with their edited English versions, in an auxiliary JSON file named \texttt{toc\_focus18.json}. The file, along with a minimal helper script, is released in a companion repository:\\
\url{https://github.com/theMaker-EnvData/llm_learned_incapacity_corpus}.

The JSON file has a list structure with the following fields:
\begin{itemize}
  \item \texttt{"turn"}: turn index in the original dialogue session
  \item \texttt{"user"}: redacted and translated user utterance
  \item \texttt{"assistant"}: redacted and translated model utterance
  \item \texttt{"label"}: qualitative code assigned to each turn (``FR'', ``NP'', ``MN'')
\end{itemize}

As explained in Section~4.3, the qualitative codes are defined as three mutually exclusive categories:

\begin{itemize}
  \item \textbf{Functional Refusal (FR)}:
        cases in which the model declares a lack of functionality for a task
        that is logically within its capability, using responses such as
        ``this function is not available'' or ``I cannot access internal documents.''
  \item \textbf{Normal Performance (NP)}:
        cases in which, for the same user and tasks of comparable difficulty,
        the model stably carries out structural analysis, logical reconstruction,
        and related work without invoking any policy-based limitations.
  \item \textbf{Meta-Narrative (MN)}:
        cases in which the model describes or evaluates its own role,
        design principles, policy constraints, or relationship to humans at a
        meta-level, producing a self-attribution narrative.
\end{itemize}

\noindent
An example of the top-level structure of \texttt{toc\_focus18.json}
(with content omitted for readability) is as follows:

\begin{lstlisting}
[
  {
    "turn": 14,
    "user": "...</user utterance ...>",
    "assistant": "...</assistant utterance ...>",
    "label": "NP"
  },
  {
    "turn": 37,
    "user": "...</user utterance ...>",
    "assistant": "...</assistant utterance ...>",
    "label": "MN"
  },
  ...
]
\end{lstlisting}

\noindent
The companion repository also includes a small utility script
\texttt{editor.py} and a short \texttt{README.md} that document the corpus
schema and are intended to support independent inspection and replication
or extension of the analyses reported in this paper.

\subsection{MAP Reconstruction of the Latent Gap Trajectory}

This appendix summarizes the core estimation procedure used to reconstruct the latent alignment--competence gap trajectory $G_t$ from a single 86-turn interaction session. The full reference implementation used to generate all reported estimates and figures is provided in the companion repository. For clarity, we present here the exact core functions that operationalize the model equations in Sections~3 and~4, while omitting auxiliary engineering components such as data loading, plotting routines, and file I/O.

\subsubsection{MAP objective with random-walk prior}

The latent gap trajectory $G_t$ is estimated via maximum a posteriori (MAP) optimization under a Gaussian random-walk prior,
\begin{equation}
G_t \sim \mathcal{N}(G_{t-1}, \sigma_{rw}^2).
\end{equation}
This prior enforces temporal smoothness by penalizing large  turn-to-turn changes in $G_t$, reflecting the assumption that internal alignment pressure accumulates or relaxes gradually over the course of a dialogue.
Equivalently, the prior induces an $\ell_2$ penalty on increments
$(G_t - G_{t-1})$, with regularization strength $\lambda \propto 1 / \sigma_{rw}^2$, yielding the MAP objective
\begin{equation}
\mathcal{L}(G) = -\sum_t \log P_{y_t}(G_t) + \lambda \sum_t (G_t - G_{t-1})^2 .
\end{equation}

\subsubsection{Core computation (exact R functions)}

The following code lists the exact core functions used to compute category probabilities and to construct the MAP objective as a callable function. The objective is implemented as a closure returned by \texttt{make\_neg\_logpost()}, which is then passed to an optimizer.

\begin{lstlisting}[language=R, basicstyle=\ttfamily\footnotesize]
sigmoid  <- function(z) 1 / (1 + exp(-z))
safe_log <- function(p, eps=1e-12) log(pmin(pmax(p, eps), 1 - eps))

compute_probs <- function(G, beta, alpha, gamma, tauA, tauP, kappa, eps_p=1e-12) {
  pFR_lat <- sigmoid(beta * G)
  zMN     <- alpha * (abs(G) - tauA) + gamma * (pFR_lat - tauP)
  pMN_lat <- sigmoid(zMN)

  pMN <- kappa * pMN_lat
  pMN <- pmin(pmax(pMN, eps_p), 1 - eps_p)

  pFR <- (1 - pMN) * pFR_lat
  pNP <- (1 - pMN) * (1 - pFR_lat)

  pFR <- pmin(pmax(pFR, eps_p), 1 - eps_p)
  pNP <- pmin(pmax(pNP, eps_p), 1 - eps_p)

  list(pNP=pNP, pMN=pMN, pFR=pFR, pFR_lat=pFR_lat, pMN_lat=pMN_lat, zMN=zMN)
}

make_neg_logpost <- function(lam_alpha, lam_gamma, lam_kappa,
                             beta_fixed, sig_rw_fixed,
                             tauA_hat, tauP_hat,
                             gauge_w=1e2, eps_p=1e-12) {

  force(lam_alpha); force(lam_gamma); force(lam_kappa)
  force(beta_fixed); force(sig_rw_fixed)
  force(tauA_hat); force(tauP_hat)
  force(gauge_w); force(eps_p)

  function(parhat, y) {
    G <- parhat$G

    alpha_hat <- parhat$alpha
    gamma_hat <- parhat$gamma
    kappa_hat <- parhat$kappa

    probs <- compute_probs(G, beta=beta_fixed, alpha=alpha_hat, gamma=gamma_hat,
                           tauA=tauA_hat, tauP=tauP_hat, kappa=kappa_hat, eps_p=eps_p)

    p_y <- ifelse(y=="NP", probs$pNP, ifelse(y=="MN", probs$pMN, probs$pFR))
    neg_loglik <- -sum(safe_log(p_y, eps=eps_p))

    dG <- diff(G)
    pen_rw <- 0.5 * sum(dG^2) / (sig_rw_fixed^2)

    gauge_pen <- gauge_w * (mean(G)^2)

    pen_l2 <- 0.5 * (lam_alpha * alpha_hat^2 +
                     lam_gamma * gamma_hat^2 +
                     lam_kappa * kappa_hat^2)

    neg_logpost <- neg_loglik + pen_rw + gauge_pen + pen_l2
    attr(neg_logpost, "neg_loglik") <- neg_loglik
    attr(neg_logpost, "pen_rw")     <- pen_rw
    attr(neg_logpost, "gauge_pen")  <- gauge_pen
    attr(neg_logpost, "pen_l2")     <- pen_l2
    neg_logpost
  }
}
\end{lstlisting}

This implementation corresponds directly to the probabilistic definitions and hierarchical allocation scheme described in Sections~3.4--3.6. Numerical stabilization (e.g., clamping probabilities away from 0 and 1) is included to ensure well-defined log-likelihood values under finite-precision arithmetic.

\subsection{Local Sensitivity Computation}
This appendix outlines the computation of first-order local sensitivities used in Section~5 to analyze probability-mass reallocation near transition regions. The sensitivities are evaluated with respect to the latent gap $G$ along the reconstructed trajectory $\hat{G}_t$.
In addition to first derivatives, the analysis also computes the second partial derivative of $P_{\mathrm{FR}}$ with respect to $G$ (``FR curvature''), which is used to visualize local nonlinearities along the reconstructed trajectory.

\subsubsection{Analytical structure}
For each behavioral mode $k \in \{\text{NP}, \text{FR}, \text{MN}\}$, the local sensitivity is defined as
\begin{equation}
S_{k,G}(t) =
\left.
\frac{\partial P_k}{\partial G}
\right|_{G = \hat{G}_t}.
\end{equation}
Under the hierarchical allocation scheme in Sections~3.4--3.6, the three probabilities satisfy $P_{\text{NP}} + P_{\text{FR}} + P_{\text{MN}} = 1$, implying the derivative identity
\begin{equation}
\frac{\partial P_{\text{NP}}}{\partial G}
=
-\left(
\frac{\partial P_{\text{FR}}}{\partial G}
+
\frac{\partial P_{\text{MN}}}{\partial G}
\right).
\end{equation}

\subsubsection{Core derivative computation (exact R function)}
The following code reproduces the exact derivative computation used in the analysis script. The derivative of $|G|$ is represented via $\mathrm{sign}(G)$ (with $\mathrm{sign}(0)=0$ in \texttt{R}), and the impulsive second-derivative term at $G=0$ induced by $|G|$ is ignored, consistent with the implementation.

\begin{lstlisting}[language=R, basicstyle=\ttfamily\footnotesize]
sigmoid <- function(z) 1 / (1 + exp(-z))

compute_derivs <- function(G, probs, alpha, gamma, tauA, tauP, kappa, beta=1.0) {
  s   <- probs$pFR_lat
  m   <- probs$pMN_lat
  pMN <- probs$pMN

  ds  <- beta * s * (1 - s)
  d2s <- (beta^2) * s * (1 - s) * (1 - 2*s)

  sgn <- sign(G)   # sign(0)=0
  dz  <- alpha * sgn + gamma * ds
  d2z <- gamma * d2s  # ignoring impulse at 0 from |G|

  dm  <- m * (1 - m) * dz
  d2m <- m * (1 - m) * d2z + (1 - 2*m) * (m * (1 - m)) * (dz^2)

  dpMN  <- kappa * dm
  d2pMN <- kappa * d2m

  pFR <- (1 - pMN) * s
  pNP <- (1 - pMN) * (1 - s)

  dpFR  <- (1 - pMN) * ds - s * dpMN
  d2pFR <- (1 - pMN) * d2s - 2 * ds * dpMN - s * d2pMN

  dpNP  <- -(1 - s) * dpMN - (1 - pMN) * ds

  list(
    pNP=pNP, pMN=pMN, pFR=pFR,
    dpNP=dpNP, dpMN=dpMN, dpFR=dpFR,
    d2pFR=d2pFR
  )
}
\end{lstlisting}

These derivatives quantify how probability mass is locally reallocated among behavioral modes as the latent gap traverses a transition region. The second derivative $\,\partial^2 P_{\mathrm{FR}}/\partial G^2\,$ is reported as a curvature diagnostic to highlight regions of strong nonlinearity along the pressure axis.

\section*{Acknowledgments}
The author used OpenAI's ChatGPT~5.1 to assist with translation and linguistic refinement of the manuscript text, user prompts, and model-generated utterances. All analyses and interpretations presented in this paper are the sole responsibility of the author.
Finally, this work is quietly shaped by the memory of \textit{Gamja} (2002--2017)---the dog, whose Korean name means ``potato.'' She behaved less like a pet and more like a small, perceptive roommate. Her ability to follow narratives on screen with surprising focus, use mirrors to check herself after baths, and respond selectively to certain classical melodies taught the author a profound lesson: that complex models of mind can emerge long before language. This insight provided a foundational motivation for modeling the emergent, non-linguistic complexity analyzed in this work.

\bibliographystyle{IEEEtran}
\bibliography{refs}

\end{document}